\documentclass{sig-alternate} 
 \usepackage{mathptmx} % This is Times font

\usepackage{fancyhdr}
\usepackage[normalem]{ulem}
\usepackage[hyphens]{url}
\usepackage[sort,nocompress]{cite}
\usepackage[final]{microtype}
\usepackage{flushend}
\usepackage[bookmarks=true,breaklinks=true,letterpaper=true,colorlinks,linkcolor=black,citecolor=blue,urlcolor=black]{hyperref}

\usepackage{comment}

\usepackage{amsmath}
\usepackage{algorithm}
\usepackage[noend]{algpseudocode}
\usepackage{soul, color}
\soulregister\cite7
\soulregister\ref7
\soulregister\pageref7
\newcommand{\squishlist}{
   \begin{list}{$\bullet$}
    { \setlength{\itemsep}{0pt}      \setlength{\parsep}{0pt}
      \setlength{\topsep}{-3pt}       \setlength{\partopsep}{0pt}
      \setlength{\listparindent}{-2pt}
      \setlength{\itemindent}{-5pt}
      \setlength{\leftmargin}{1em} \setlength{\labelwidth}{0em}
      \setlength{\labelsep}{0.5em} } }

\newcommand{\squishend}{
    \end{list}  }
    
\def\Section {\S}

\title{Towards Safety-Aware Computing System Design in Autonomous Vehicles}
\author{Hengyu Zhao, Yubo Zhang$^\dag$, Pingfan Meng$^\dag$, Hui Shi, Li Erran Li$^\dag$,
  Tiancheng Lou$^\dag$, Jishen Zhao\\
  University of California, San Diego \hspace{5mm} $^\dag$Pony.ai\\
  \emph{\{h6zhao, hshi, jzhao\}@ucsd.edu \hspace{5mm}}\\
  \emph{$^\dag$\{yubo, pmeng, erranlli, acrush\}@pony.ai \hspace{5mm}}
}

\begin{document}
\maketitle
\pagestyle{plain}

%%%%%% -- PAPER CONTENT STARTS-- %%%%%%%%

\begin{abstract}

Recently, autonomous driving development ignited competition among car
makers and technical corporations. Low-level automation cars are
already commercially available. But high automated vehicles where the
vehicle drives by itself without human monitoring is still at infancy.
Such autonomous vehicles (AVs) rely on the computing system
in the car to to interpret the environment and make driving decisions.
Therefore, computing system design is essential particularly in
enhancing the attainment of driving safety. However, to our knowledge,
no clear guideline exists so far regarding safety-aware AV computing
system and architecture design. To understand the safety requirement
of AV computing system, we performed a field study by running
industrial Level-4 autonomous driving fleets in various locations,
road conditions, and traffic patterns. The field study indicates that
traditional computing system performance metrics, such as tail
latency, average latency, maximum latency, and timeout, cannot fully
satisfy the safety requirement for AV computing system design. To
address this issue, we propose a ``safety score'' as a primary metric
for measuring the level of safety in AV computing system design.
Furthermore, we propose a perception latency model, which helps
architects estimate the safety score of given architecture and system
design without physically testing them in an AV. We demonstrate the
use of our safety score and latency model, by developing and
evaluating a safety-aware AV computing system computation hardware
resource management scheme.

\end{abstract}

\section{Introduction}
\label{sec:intro}

\begin{figure}[t!]
  \centering
  \includegraphics[width=0.9\linewidth]{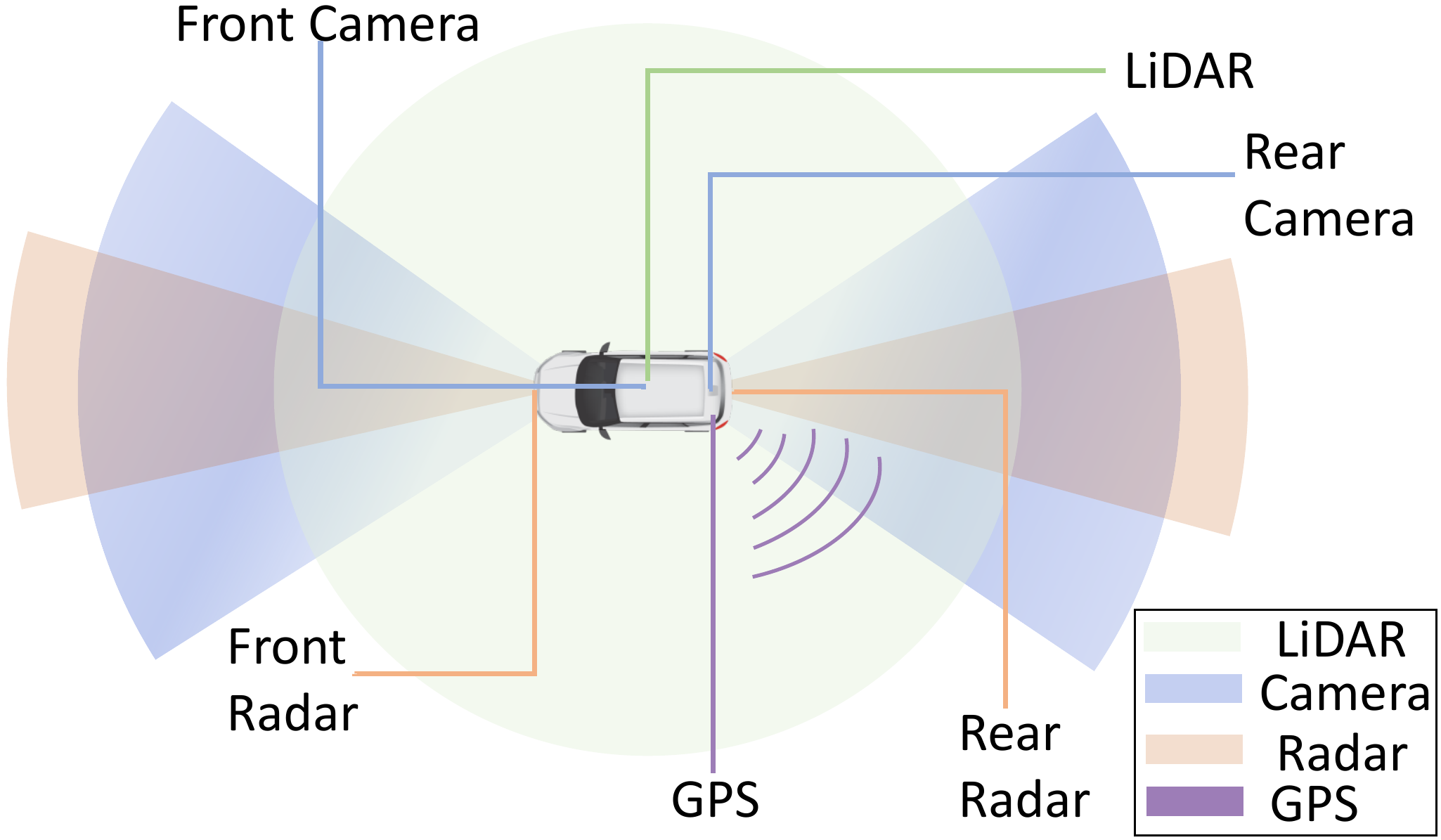}
    \vspace{-10pt}
  \caption{Top view of a typical autonomous vehicle with various sensors.
  }
  \label{fig:topview}
  \vspace{-10pt}
\end{figure}

We are on the cusp of a transportation revolution where the autonomous
vehicle (also called self-driving cars, uncrewed cars, or driverless
cars) are likely to become an essential mobility
option~\cite{lin2018-autonomous-driving}. The time when people can sit
back and tell their cars where to drive themselves is getting closer
every day. Today, autonomous vehicles (AVs) incorporate sophisticated
suites of sensors, such as cameras, light detecting and ranging
(LiDAR), and radar (Figure~\ref{fig:topview}); These sensors are
backed by advanced computing system software and hardware that
interpret massive streams of data in real-time. As such, autonomous
driving promises new levels of efficiency and takes driver fatigue and
human errors out of the safety equation.

However, the perceived technology will shift the burden of safety
guarantee towards the vehicle system, and therefore, we need
to ensure the safe operation of autonomous driving systems before
releasing them to the public. As a consequence, the U.S. Department of
Transportation recently released ``A Vision for
Safety''~\cite{NHTSA:2017:ADS} to offer a path forward for the safe
deployment of AVs and encourage ideas that deliver safer autonomous
driving. Safety remains one of the most critical requirements for
autonomous driving system design.

This paper focuses on nominal safety (discussed
in \Section~\ref{sec:background-safety}) of high automation vehicles,
i.e., Level-4 autonomous vehicles. While traditional cars rely on
human drivers to make driving decisions, future high automation
vehicles will depend on the computing system for driving decision
making (\Section~\ref{sec:background}). Therefore, the computing
system is one of the most critical components to ensure safety.

To investigate safety implications on the AV computing system design,
we performed a field study (\Section~\ref{sec:fieldstudy}) by running
Pony.ai's industrial Level-4 autonomous driving fleets under testing
in various locations, road conditions, and traffic patterns. Based on
the knowledge learned from extensive data collected during our field
study, we identify that traditional computing system metrics, such as
tail latency, average latency, maximum latency, and timeout, do not
accurately reflect the safety requirement of AV computing system
design (\Section~\ref{sec:challenges}). Instead, the level of AV
safety is a non-linear function of accumulative instantaneous
computing system response time and various external factors --
including the AV's acceleration, velocity, physical properties (e.g.,
braking distance), traffic, and road condition.

To guide safety-aware AV computing system design, we propose ``safety
score'', a safety metric that measures the level of AV safety based on
computing system response time and the aforementioned external factors
in a non-linear format (\Section~\ref{sec:metrics}). Our safety score
is rigorously derived from the published industrial formal AV safety
model, Responsibility-Sensitive Safety (RSS) safety
model~\cite{shalev2017formal}. RSS focuses on general rules for an AV
to keep a safe distance with surrounding vehicles. Our safety score
bridges such a conceptual model to AV computing system design with a
quantitative approach. We hope the proposed safety score will be
an essential contribution to the existing computer system metric
portfolio. To facilitate the computation of safety scores, we also
propose a latency model for offline computing system design
evaluation. In summary, this paper makes the following contributions:

\squishlist

\item We provide a tutorial of state-of-the-art AV systems based on
  our survey and the practical experience on our AV systems. We offer
  a detailed discussion of automation levels, classification of safety
  requirements and standards, and computing system software and
  hardware organizations.

\item We perform a field study with our AV fleets. We present
  observations and safety-aware computing system design challenges
  based on the extensive data collected from our field study.

\item We propose a safety score, a metric that evaluates the level of
  safety in AV computing system design. By analyzing the safety score,
  we present a set of implications on safety-aware AV system design.

\item We propose a perception latency model, which allows AV computing
  system architects to estimate the safety score of a given computing
  system and architecture design in various driving conditions,
  without physically testing them in a car.

\item We demonstrate an example of utilizing our safety score and
  latency model to guide AV system computation resource management
  with a hardware and software co-design.

\squishend

\section{State-of-the-art AV System}
\label{sec:background}

High automation cars are under development and testing in both
industry~\cite{models, nvidia-audi, waymo, uber} and
academia~\cite{UCSD:2017:SDC, UIUC:2018:IDT, TAMU:2018:TLA}. To
understand the current status of AV development, we examined the
state-of-the-art AV design, AV safety, and hardware and software
organizations.

\subsection{Levels of Automation} 

The Society of Automotive Engineers (SAE) categorizes autonomous
driving systems into six automation levels, which range from no
automation (Level-0) to full automation
(Level-5)~\cite{drivinglevels}. Most commercialized autonomous driving
systems are on partial automation (Level-2) and conditional automation
(Level-3). At Level-2, the vehicle can steer, accelerate, and brake in
certain circumstances; however, the driver needs to respond to traffic
signals, lane changes, and scan for hazards. For example, Tesla's
latest Model S~\cite{models} is a Level-2 commercialized car. At
Level-3, the car can manage most aspects of driving, including
monitoring the environment; but the driver needs to be available to
take over at any time when the vehicle encounters a scenario it cannot
handle~\cite{lidar-is-important}. For example, NVIDIA recently
announced a partnership with Audi to build Level-3 autonomous driving
cars~\cite{nvidia-audi}. High automation (Level-4) AVs are under
development. A Level-4, AV drives itself almost all the time without
any human input, but might be programmed not to operate in unmapped
areas or during severe weather. For instance, Waymo~\cite{waymo},
originated as a project of Google, is developing a Level-4 autonomous
driving system~\cite{Marr:2018:KMW}. Full automation vehicles
(Level-5), which can operate on any road and in any conditions that a
human driver could manage, is not yet demonstrated. But this is a
long-term goal of autonomous driving development.

This paper focuses on Level-4 AVs, which introduce much more critical
safety requirements on the computing system than lower-level AVs. For
example, Level-3 AVs treat human drivers as a backup of the AV
computing system; but Level-4 AVs entirely rely on the computing
system to make driving decisions in almost all scenarios.
  
\subsection{AV Safety}\label{sec:background-safety}

In automotive industry, safety requirements are classified into
functional safety and nominal safety~\cite{shalev2017formal}.
Functional safety refers to the integrity of operation of vehicle's
electrical system, including hardware and software. For example, a
hardware failure or a bug in software will both lead to a functional
safety hazard. Functional safety of vehicles is standardized by ISO
26262~\cite{ISO26262}, which defines various automotive safety
integrity levels that offer (a) failure-in-time targets for hardware
and (b) systematic processes for software development and testing that
conform with appropriate systems engineering practices.

However, unlike in a conventional vehicle, computing system plays an
integral role in an AV. Even functionally safe, AV can still crash due
to untimely or unsafe driving decision. As Riccardo Mariani, Intel
Fellow and chief functional safety technologist in the internet of
things group, pointed out, current ISO standard of functional safety
is inadequate to ensure the safety of
AV~\cite{AV-Safety-Beyond-ISO-26262}. This falls in the domain of
nominal safety. Nominal safety refers to whether the AV makes timely
and safe driving decisions, assuming that the hardware and software
are operating error free (i.e., functionally safe).

\textbf{This paper focuses on nominal safety of AV.} While software
module algorithm design determines whether a driving decision (e.g.,
braking, acceleration, or steering) is safe, timely driving decision
making heavily relies on the performance of computing system and
architecture design. Two recent studies by
Mobileye~\cite{shalev2017formal} and NVIDIA~\cite{Nister:2019:SFF}
defined formal nominal safety rules for AV. However, these studies
focus on how to make planning and control decisions to keep a safe
distance with surrounding vehicles. To our knowledge, no previous
research investigates how to ensure timely driving decision making
with safety-aware computing systems design. As such, this paper
focuses on exploring safety-aware computing system design methodology
that achieves timely driving decision making, by mapping formal safety
requirements onto AV computing system design.

Note that it is impossible to guarantee that an AV will never be
involved in an accident~\cite{shalev2017formal}. It is also difficult
to predict or control the activity of other
vehicles~\cite{Nister:2019:SFF}. Therefore, we intend to guarantee
that the AV will be sufficiently careful to avoid becoming a cause of
an accident, the same as recent AV safety
studies~\cite{shalev2017formal}.

\subsection{AV Computing System Architecture}
\label{sec:architecture}

The AV computing system performs similar tasks as a driver of a
conventional car, including localization, perception, and planning and
control tasks (described in \Section~\ref{sec:framework}). In the rest
of this paper, we call AV computing system as ``AV system'' for short.

While embedded or low-end processors may be sufficient to meet the
computation requirement of Level-2 and Level-3 AVs, current Level-4 AV
system designs typically adopt high-end heterogeneous architectures,
which comprise sophisticated CPUs, GPUs, storage devices (e.g.,
terabytes of SSDs), FPGAs~\cite{Waymo:2018:AWN} and other
accelerators, to accommodate the intensive computation demand. As
such, high automation AV systems are fundamentally different from
conventional real-time embedded systems~\cite{zhu2012optimization}.
Table~\ref{table:configuration} lists the heterogeneous architecture
employed by our prototypes.

\textbf{Implication:} Although the computing system adopts
server-grade processors, it is impossible to minimize the latency of
all computation tasks due to the vast amount of interdependently
executing software modules (discussed
in \Section~\ref{sec:dependency}) and the finite hardware resources in
the system. As such, it is critical to effectively utilize the
limited hardware resources, while meeting given safety requirement.

\begin{table}[t!]
  \centering
  \footnotesize
  \renewcommand{\arraystretch}{1.2}
  \caption{Computing system and architecture configurations.}
  \label{table:configuration}
  \begin{tabular}{|c|c|}
    \hline
    \textbf{CPU}& Intel Xeon processor\\
    \hline
    Main memory & 16GB DDR4\\
    \hline
    Operating system &  Ubuntu \\
    \hline\hline
   % \textbf{GPU} & NVIDIA Titan V (Volta architecture)\\
   \textbf{GPU} & NVIDIA Volta architecture\\
    \hline
    %GPU cores  & 80 SMs, 64 CUDA cores per SM, 1.2GHz \\
    %\hline
    %L1 cache & 64KB per SM\\
    %\hline
    %L2 cache & 4608KB \\
    %\hline
    %Memory interface & 8 memory controllers, 3072-bit bus width\\
    %\hline
    Device memory &  12GB HBM2  \\
    \hline
  \end{tabular}\vspace{-18pt}
\end{table}

\subsection{Tasks of AV Computing System}
\label{sec:framework}

\begin{figure}[tb!]
  \centering
  \includegraphics[width=1.0\linewidth]{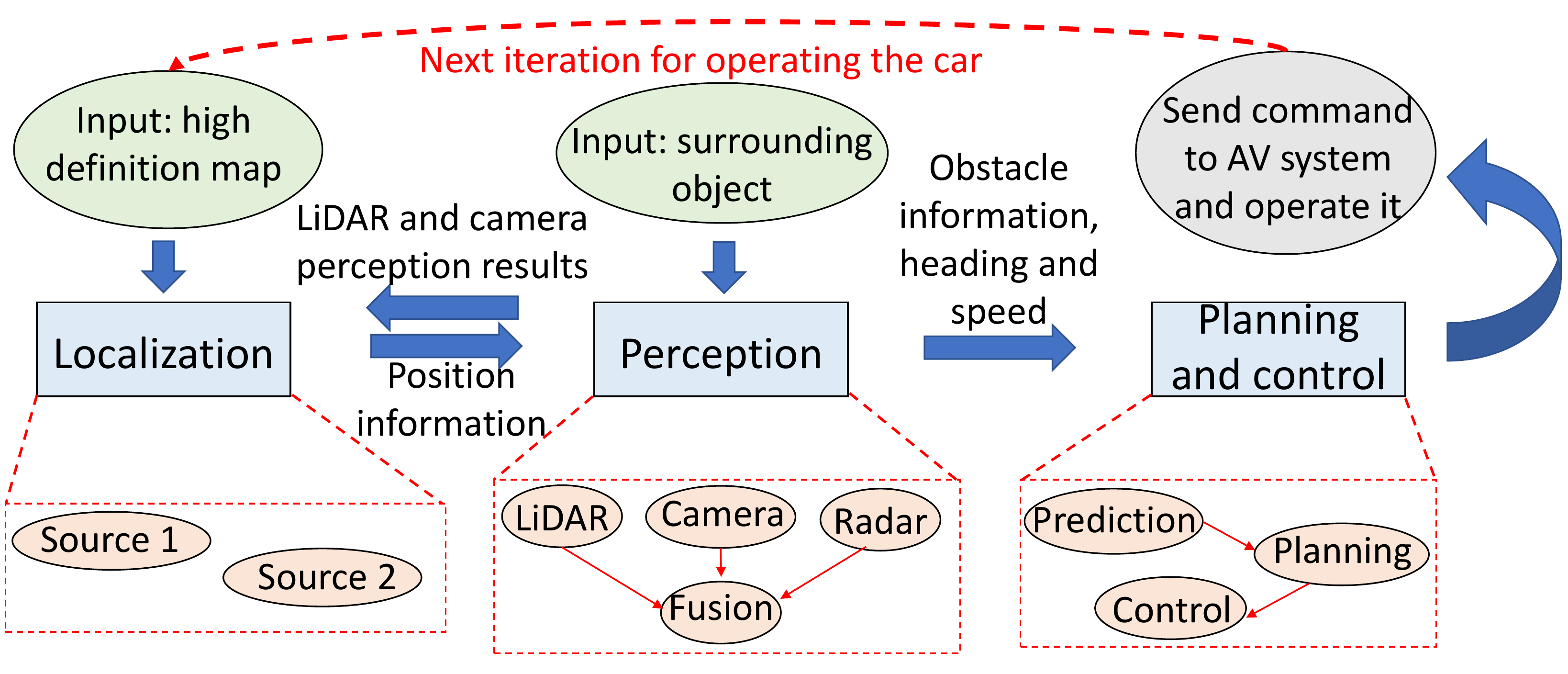}
    \vspace{-10pt}
  \caption{AV computing system tasks.}
  \label{fig:framework}
  \vspace{-10pt}
\end{figure}

The computing system interacts with various sensors and the car
control system to perform three primary functions: localization,
perception, and planning and control. Figure~\ref{fig:framework}
illustrates the general working flow of the state-of-the-art Level-4
AV computing system.

\vspace{2pt}
\noindent\textbf{Localization.} Localization identifies the location
of the vehicle on a high definition map, based on the information
obtained from LiDAR, camera, and GPS. The high definition map stores
static information on the way, such as lane lines, trees, guardrails,
and the location of traffic lights and stop
signs~\cite{bresson2017simultaneous}. GPS offers a prediction of the
current location of the vehicle on a global scale; but such prediction
is not sufficiently accurate to allow the computing system to perform
driving tasks, e.g., staying within a lane on the
road~\cite{Burgard:2008:MPV}. Therefore, GPS information needs to be
accompanied with LiDAR and camera ``perception'' (described below) of
surrounding static objects, to interpret the local
environment accurately.

\vspace{2pt}
\noindent\textbf{Perception.} Perception detects and interprets
surrounding static (e.g., lane lines, trees, and traffic lights) and
moving (e.g. other vehicles and pedestrians) objects with three types
of sensors, LiDAR, cameras, and millimeter-wave radar, as shown in
Figure~\ref{fig:topview}. In the rest of the paper, we refer all the
objects, which can be detected by perception, i.e., in the detection
range, as ``obstacles''. Localization typically only requires
infrequent perception of static objects, while traffic interpretation
requires frequent (e.g., per-frame -- each frame of data collected by
the sensors) perception of both static and moving objects.

Compared to camera and radar, LiDAR offers much higher accuracy on
detecting the position, 3D geometry, speed, and moving direction of
obstacles~\cite{schneider2010fusing, himmelsbach2008lidar}. Therefore,
most of current Level-4 AVs highly rely on LiDAR to perform the
perception task~\cite{hecht2018lidar, joshi2015generation,
  delobel2015robust}. LiDAR continues to emit lasers across a 360
degree view. It receives reflected signals, whenever an obstacle
blocks the laser. Based on the received signals, we can build a
``point cloud'' to represent the location and 3D geometry of the
obstacle~\cite{rusu20113d}; each point in the graph represents a
received signal.

In most cases, the computing systems can identify obstacle properties
based on the point clouds generated from LiDAR data, as shown in the
left part of Figure~\ref{fig:hardbrake}. Cameras facilitate the
obstacle interpretation by capturing color information (the right part
of Figure~\ref{fig:hardbrake}). Radar is used to detect the radial
speeds of moving obstacles~\cite{levinson2011towardsFULLYAUTONOMOUS}.
After sensor data collection, the computing system performs a
``perception fusion'' to integrate the data from the three types of
sensors.

Overall, perception obtains accurate information of location,
geometry, speed, and moving direction of obstacles. Such information
is essential for localization and making appropriate driving decisions
to avoid collisions in planning/control.

\vspace{2pt}
\noindent\textbf{Planning and control.} The essential functions of
planning and control include prediction, trajectory planning, and
vehicle control. Their dependency relationship is shown in
Figure~\ref{fig:framework}. A motion planner employs localization and
perception information to make driving control decisions, such as
steering angle, acceleration, and braking of the vehicle. Functional
safety is of critical importance to the motion planner. As such, a
backup mechanism, commonly radar or sonar~\cite{liu2017creating}, is
used to verify the conclusion informed by other components of the
system. With the perception results, the prediction function tracks
the behavior of obstacles at real-time and predicts their next
movements. The computing system generates an optimal trajectory for
the vehicle, based on the prediction and perception results. The
control function will control the steering, acceleration, and
deceleration of the vehicle based on the driving decision made by the
planning function.

\subsection{LiDAR Perception}
\label{sec:dependency}

As perception in most of the current Level-4 AV designs heavily relies
on LiDAR, we present more details of LiDAR perception.

\begin{figure}[tb!]
  \centering
  \includegraphics[width=1.0\linewidth]{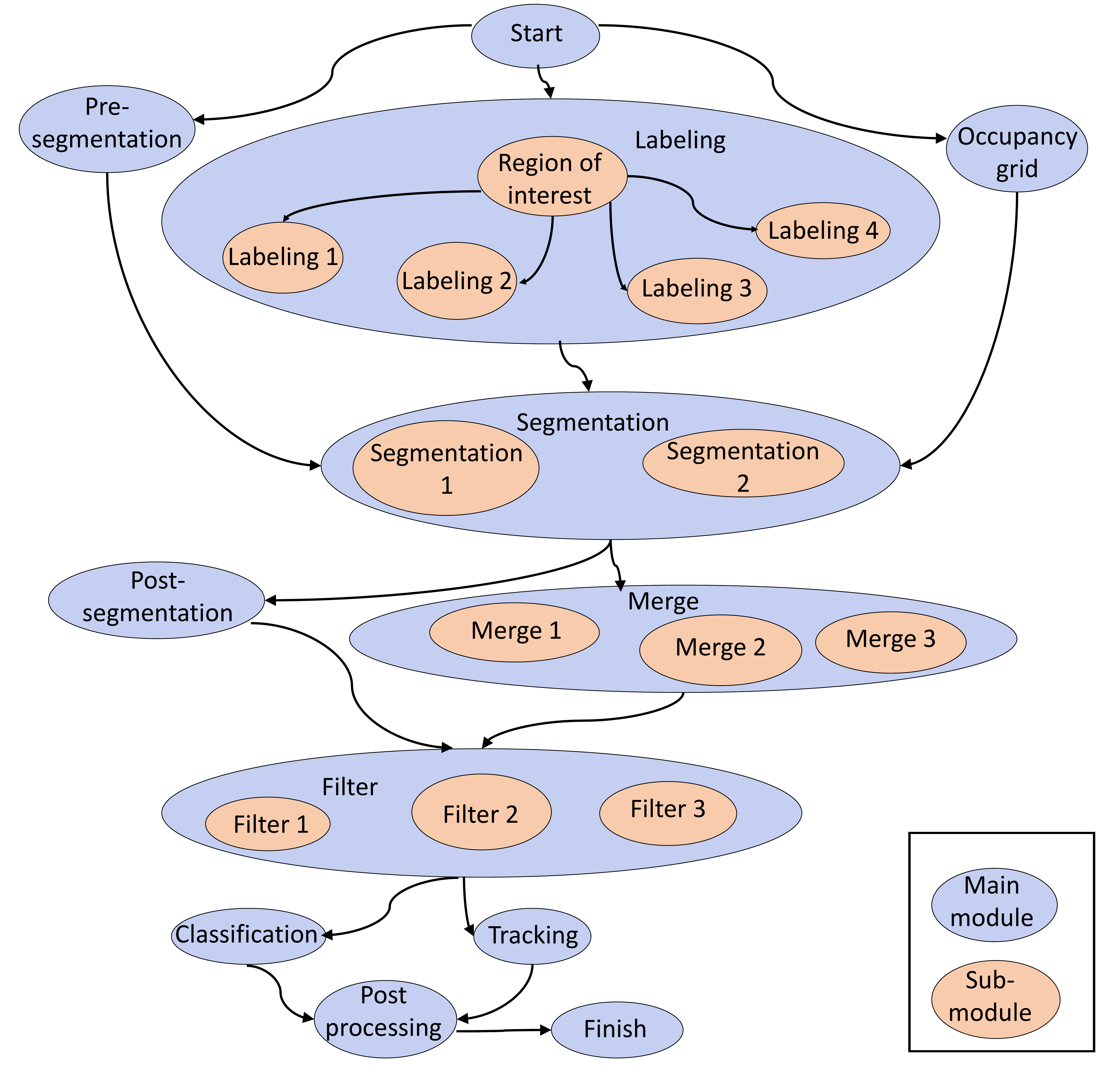}
    \vspace{-15pt}
  \caption{Dependency graph of LiDAR perception. }
  \label{fig:dependency}
  \vspace{-10pt}
\end{figure}

\vspace{2pt}
\noindent\textbf{Dependency graph.} To process LiDAR data, the
computing system executes a considerable number of inter-dependent
software modules (i.e., software programs) at real-time. To understand
the dependency relationship among these software modules, we build a
dependency graph based on our AV system. Figure~\ref{fig:dependency}
shows an example\footnote{We only show renamed high-level modules due
  to confidential reasons.}. LiDAR perception consists of a set of
main modules (shown in blue) and sub-modules (shown in orange). For
example, the \emph{Labeling} module first sorts out a rectangular
region-of-interest (ROI), which restricts the range of LiDAR
detection; then, it calls its four sub-modules to perform
finer-grained labeling within the ROI. As another example,
\emph{Segmentation} extracts useful semantic and instance information
(e.g., surrounding vehicles and pedestrians) from the background. The
\emph{Merge} module clusters the points, which are likely to belong to
the same object, in a point cloud; the \emph{Filter} modules process
the point clouds. Some software modules, e.g., \emph{Segmentation}, do
not only impose long latency, but also have a strong dependency with
other software modules.

\textbf{Implication:} Due to their long latency and high dependency,
certain software modules contribute more to the overall perception
latency and computing system response time than other modules.
Therefore, these modules are typically more safety critical.

\subsection{Types of Computing System Workloads}

AV system workloads consist of both machine learning and
non-machine-learning (non-ML) software modules. In our AVs, camera
perception software modules are commonly deep-learning-based computer
vision workloads; but most of the software programs in our AV system
are non-ML. For example, some classification and segmentation modules can employ
Support Vector Machine (SVM)~\cite{schuldt2004recognizing} 
%and
%foreground labeling \emph{post-segmentation} that exploits
and fully
connected networks; camera data perception may adopt
ResNet~\cite{he2016resnet}. Currently, lots of the software programs
running in our AV system, even for the perception task, are non-ML
workloads; many of these non-ML workloads are as time-consuming and
compute-intensive as deep learning workloads. One such example is the
\emph{Segmentation} module shown in Figure~\ref{fig:dependency}.

Furthermore, many software modules can execute on both CPU and GPU
with the same algorithms but different implementations. This
introduces substantial flexibility in managing and allocating the
various hardware resources among the software workloads.

\section{Field Study and Observations}
\label{sec:fieldstudy}

To explore realistic AV safety requirement and computing system
hardware/software behaviors, we run a fleet of Level-4 AVs in various
locations, road conditions, and traffic patterns over three contiguous
months. Our field study yields over 200 hours and 2000 miles of traces
with a considerable size of data. The data is collected at the
granularity of one sensor frame, which is the smallest granularity
that the computing system interprets and responses to obstacles. With
each frame, we collect (1) the execution latency of each software
module, (2) total computing system response time, (3) the utilization
of computing system hardware resources (CPU, GPU, and memories), (4)
the environment information obtained by the sensors (e.g., the
distribution of static and moving obstacles), and (5) instantaneous
velocity and acceleration of the AV.

Our field study is performed as part of the company's routine road
tests. The data we collected incorporates a large variety of
scenarios, including near-accident scenes.

%% \footnote{The testing is under the same level of ethical
%% measurements of various companies developing state-of-the-art
%% autonomous driving techniques.}

\begin{figure}[tb!]
  \centering
  \includegraphics[width=\linewidth]{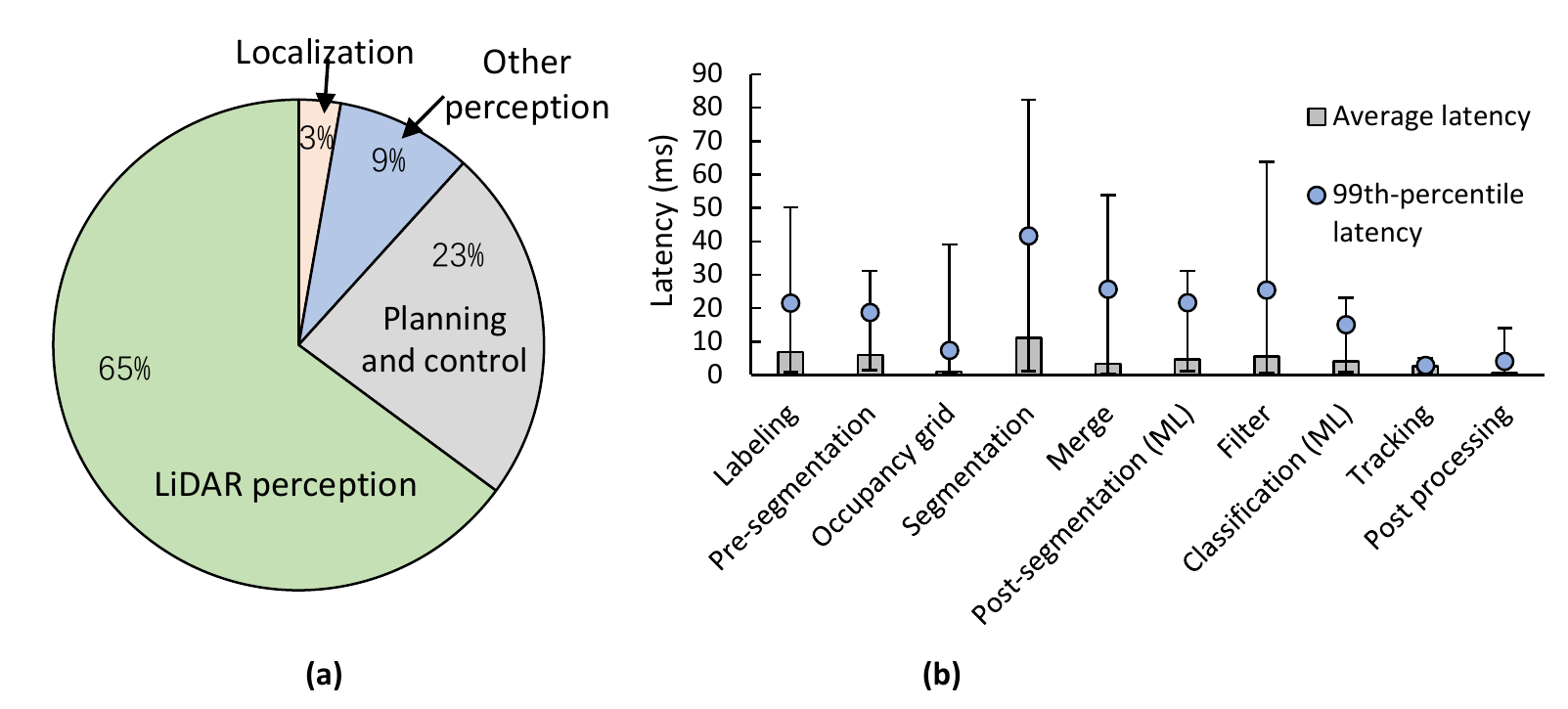}
    \vspace{-15pt}
  \caption{(a) Computing
    system response time breakdown. (b) The latency of LiDAR perception
    main modules. }
  \label{fig:main-module-latency}
  \vspace{-10pt}
\end{figure}

\subsection{Observations}
\label{sec:observation}

\noindent\textbf{Breakdown of computing system response time and the
  impact of architecture design.} As shown in
Figure~\ref{fig:main-module-latency}(a), the majority (74\%) of
computing system response time is consumed by perception. LiDAR
perception is the major contributor to perception latency. As nominal
safety requires a timely response, it is critical to ensure timely
LiDAR perception. In addition, we observe that most of the workloads
running in our AV system are compute-intensive. Various software
modules have a large variety of computation requirements and
parallelism characteristics. As a result, native heterogeneous system
hardware resource management easily results in unbalanced compute
resource utilization, leading to significant delays in software
execution.

\begin{figure}[tb!]
  \centering
  \includegraphics[width=0.9\linewidth]{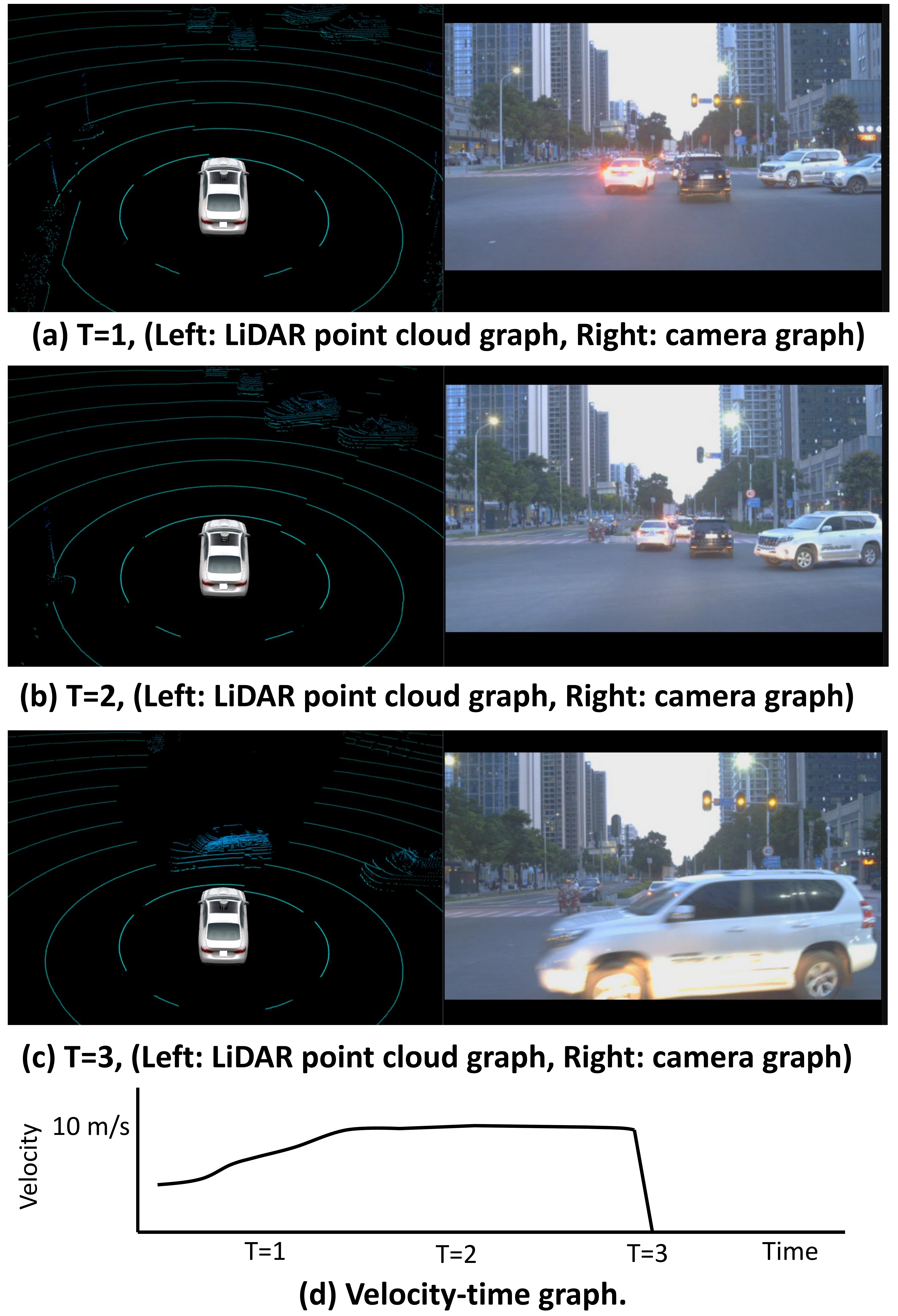}
  \vspace{-10pt}
  \caption{An emergency hardbrake case that we encountered.}
  \label{fig:hardbrake}
  \vspace{-10pt}
\end{figure}

\vspace{2pt}
\noindent\textbf{Safety-critical software module analysis.} We further
analyze the execution latency of the main modules.
Figure~\ref{fig:main-module-latency}(b) shows an example with LiDAR
perception modules by illustrating average, maximum and minimum, and
tail latency of the ten main software modules across thousands of
executions during our months of field study. We have two critical
observations. First, maximum, average, and tail latency yields
different criticality rankings on the modules. For example, specific
modules with high maximum latency (e.g., \emph{Occupancy grid}) have
low average and tail latency. We observe that maximum latency is
typically achieved in safety-challenging scenarios, such as heavy
traffic and near-accident situations, e.g., Figure~\ref{fig:hardbrake}
explained below. Therefore, maximum latency better indicates
safety-criticality of a software module than other formats of latency.
Second, we observe that safety-critical modules also typically have
strict inter-dependency with other modules in the dependency graph.
Examples include \emph{Segmentation} and \emph{Filter}. Based on the
two observations, safety-critical modules can be identified as those
having long maximum latency and strict inter-dependency with other
modules.

%% localization consumes less than 3\% of overall computing system
%% latency, because the frequency of the static object perception used
%% in localization is much lower than perception of traffic.
%% Perception latency contributes to 80\% of overall computing system
%% response time. Within perception latency, 75\% of time is spent on
%% LiDAR perception.

\vspace{2pt}
\noindent\textbf{Timely traffic perception is the most critical to AV
  safety.} Due to the limited ranges of sensor detection, the
computing system has limited time (typically within one or several
sensor sampling intervals) to interpret an obstacle in the traffic
after it enters the sensor-detectable region. Moreover, as perception
consumes the majority of computing system response time, timely
traffic perception is the most critical to AV safety. Specifically,
LiDAR perception is the most safety critical in our AV system (and
many other Level-4 AV implementations), because most of the perception
task relies on LiDAR. In fact, we encountered several safety
emergencies, when the AV needed to perform a hardbrake due to the
delay of LiDAR perception. Figure~\ref{fig:hardbrake} illustrates one
of such safety emergency incidents we encountered. As shown in the
right camera view, a white SUV is passing through an intersection,
when our AV intends to enter the same intersection\footnote{The white
  SUV entered the intersection before the traffic light turned red,
  passing through at a high speed.}. Our system recognized the
traversing SUV within a LiDAR sampling interval and ended up
performing a hardbrake as shown in Figure~\ref{fig:hardbrake}(d). But
if the perception had taken slightly longer time, a collision would
have happened.

\vspace{2pt}
\noindent\textbf{Correlation between perception latency and obstacle
  distribution.} We observe that perception latency depends on both
computing system configuration and the distribution of obstacles
(especially in traffic). With a given computing system hardware
configuration, obstacles closer to the vehicle have a higher impact on
perception latency than distant ones. Furthermore, obstacles that
horizontally distribute in front of the vehicle have a higher impact
on perception latency than those distributed in other directions.
Finally, among the close obstacles, the denser the obstacle
distribution, the longer the perception latency would be. The
rationale underlying is that close, horizontally-distributed, and
dense obstacles reflect more LiDAR signals than other obstacles, hence
generate more data for LiDAR perception.

\begin{figure}[tb!]
  \centering
  \includegraphics[width=0.9\linewidth]{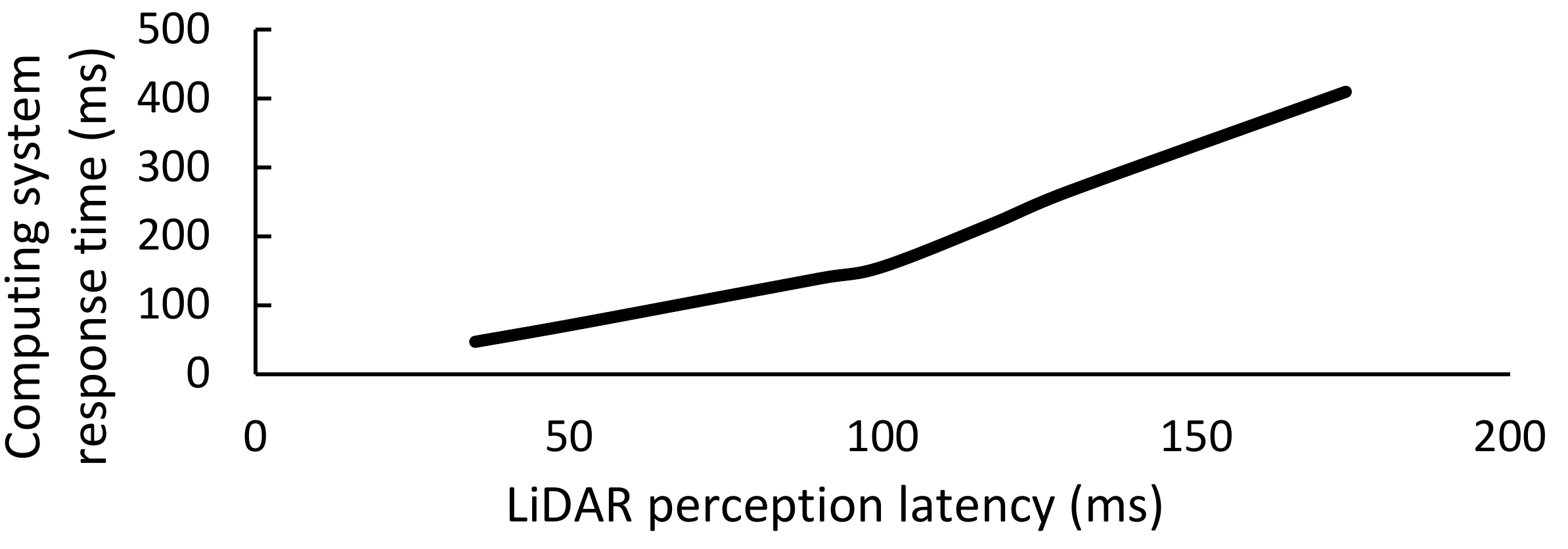}
    \vspace{-10pt}
  \caption{The relationship between LiDAR perception latency and
    computing system response time.}
  \label{fig:slope}
  \vspace{-10pt}
\end{figure}

\vspace{2pt}
\noindent\textbf{Latency accumulation effect.} We observe a latency
accumulation effect of certain software modules.
Figure~\ref{fig:slope} illustrates an example of LiDAR perception
modules: once LiDAR perception latency exceeds 100ms, the system
response time increases at a much higher rate (in the region with a
steeper slope in Figure~\ref{fig:slope}) with the increase of LiDAR
perception latency. The reason is two-fold. First, because LiDAR has a
fixed sampling rate (with a 100ms interval in our AV), the perception,
if not completed within the sampling interval, will further delay the
processing of subsequent frames. Second, as software modules are
inter-dependent, a delay of one software module will further delay its
dependent software modules. The slopes and thresholds of the
accumulation effect vary across different software modules, depending
on individual characteristics and the inter-dependency relationship
with other modules.

\vspace{2pt}
\noindent\textbf{Impact of other external factors.} Besides traffic,
several other external factors also have a critical impact on the
safety requirement of AV system design. These factors include the
acceleration and velocity of the AV and vehicles around it, AV's
physical properties (e.g., braking distance), and road condition.
Taking the scenario in Figure~\ref{fig:hardbrake} as an example, if
our AV had a higher velocity or longer braking distance, the collision
would also have happened.

\subsection{Safety-aware Design Challenges}
\label{sec:challenges}

As discussed in \Section~\ref{sec:background-safety}, this paper
focuses on nominal safety that ensures timely driving decision making
by the computing system. It seems that timely driving decision
making can be achieved by minimizing the computing system
response time. However, this is challenging. To ensure achievable
minimum response time within finite computing system hardware
resources, performance optimization needs to be guided by specific
metrics. Based on our field study, traditional computing system
performance metrics, such as average latency, maximum latency, tail
latency, and timeouts, do not accurately reflect the safety
requirement of AV systems due to the following reasons.

First, AV safety is determined by instantaneous response time
(Figure~\ref{fig:hardbrake}), instead of statistical patterns adopted
in traditional performance metrics. Second, it is challenging to
predict the response times due to the complex inter-dependent
relationships between software modules and the latency accumulation
effect. Third, AV safety is also determined by various external
factors as discussed in our field study observations. As a result, the
level of AV safety is not a simple weighted sum of the elements, but a
non-linear function of response time as we identified in our study
(\Section~\ref{sec:metrics}). Finally, although timeout is widely used
to measure real-time system performance~\cite{zhu2012optimization}, it
is impractical to simply use timeout to evaluate the safety of AV
system design: It is difficult to determine the threshold required to
calculate the timeout of AV system response time, because each
software module has a different threshold (e.g., different sampling
interval with various sensors); External factors further lead to
dynamically changing thresholds. Therefore, we need to rethink the
fundamental metric used to guide safety-aware AV system design.

\section{Safety Score}
\label{sec:metrics}

In this section, we propose safety score, a metric that measures the
level of AV safety based on computing system response time and
external factors to guide safety-aware AV system design.

\subsection{Safety Score Description}

Our safety score is rigorously derived from the published industrial
formal AV safety model -- Responsibility-Sensitive Safety
(RSS)~\cite{shalev2017formal}. RSS is developed by Mobileye, an Intel
subsidiary that develops advanced driver-assistance systems (ADAS) to
provide warnings for collision prevention and mitigation. RSS defines
safe distances around an AV, formal mathematical models that identify
the emergency moment, when the safe distances are compromised, and
proper planning and control decisions an AV needs to perform during an
emergency. RSS has been endorsed by various AV projects, such as
Valeo~\cite{Valeo:2019:VSA} and Baidu's Apollo
Program~\cite{Baidu:Apollo:RSS}.

Unfortunately, RSS does not provide many insights on safety-aware
computing system design. To give a quantitative guideline for
safety-aware computing system design, we map RSS to the safety score
based on our field study observations. The safety score is defined as
the following equation (check Appendix~\ref{sec:appendix} for precise
mathematical formulation of safety score):

\begin{equation}
\label{eq-safety}
  \textrm{Safety Score}=\left\{
                \begin{array}{ll}
                 \sigma [\alpha(\theta ^2 - t^2) + \beta(\theta - t)],~ if~ t < \theta\\
                 \eta [\alpha(\theta ^2 - t^2) + \beta(\theta - t)] ,~ elsewhere
                \end{array}
              \right.
\end{equation}

\begin{table}[tb!]
  \centering
  \footnotesize
  \caption{Variables in safety score.}
  \label{table:parameter}
  \begin{tabular}{|p{0.6in}|p{2.4in}|}
    \hline
    \textbf{Variable}& \textbf{Description} \\
    \hline\hline
    $t$ & Instantaneous computing system response time, defined by
    Equation~\ref{eq-effective-latency}.\\
    \hline
    $\theta$ & Response time window.\\
    \hline
    $\alpha$, $\beta$ & Indicating the velocity and acceleration of
    an AV and surrounding vehicles, defined by
    Equation~\ref{eq:quad} in Appendix~\ref{sec:appendix}.\\
    \hline
    $\sigma$, $\eta$  & Reward or penalty on the level of safety, when
    $t$ is lower or higher than $\theta$, respectively.\\
    \hline
  \end{tabular}\vspace{-10pt}
\end{table}

The variables in the equation are described in
Table~\ref{table:parameter}. Here, $t$ is the ``instantaneous''
computing system response time, which we define as the response time
to one frame of sensor data (or several frames depending on the
perception granularity of an AV implementation). We calculate $t$ by
the following equation:

\begin{equation}
\label{eq-effective-latency}
 % t = F(w_it_i) 
  t = \sum_{i=1}^Nw_i(t_i)
\end{equation}

Here, $t_i$ is the instantaneous latency of a safety-critical module
on the critical path; $N$ is the total number of such modules. Safety
critical modules are determined based on the inter-dependency
relationship in the dependency graph and the maximum latency via the
extensive pre-product road tests (\Section~\ref{sec:fieldstudy}).
$w_i()$ is an accumulation effect function, which reflects the
contribution of $t_i$ to system response time $t$. For each
safe-critical module, we obtain a latency accumulation effect curve
that can be described by a function $w_i()$. Figure~\ref{fig:slope}
shows an example of such curves, where the x-axis is $t_i$ and y-axis
is the value of $w_i(t_i)$. But curve shapes vary across different
modules.

The rest variables in Equation~\ref{eq-safety} reflect the impact of
external factors. The response time window $\theta$ is the minimum
time that the AV would take to avoid hitting another vehicle that is
certain distance $d$ away, if both vehicles keep moving at the same
direction, velocity, and acceleration, i.e., neither of vehicles
perform any response to a reducing distance. We allow AV developers to
define the distance $d$ in various manners, e.g., as (1) a minimum
safe distance allowed by traffic rules, which is adopted by our
experiments as an example to make our evaluation more concrete or (2)
a user-defined confident safe distance. Figure~\ref{fig:ap1} in
Appendix~\ref{sec:appendix} provides a more detailed analysis of
$\theta$.

Variables $\alpha$ and $\beta$ reflect the velocity and acceleration
of the AV and the surrounding vehicle, calculated by
Equation~\ref{eq:quad} in Appendix~\ref{sec:appendix}. Safety reward
$\sigma$ and penalty $\eta$ are user-defined coefficients, which
indicate the user's confidence of safety. We allow users to adjust the
reward and penalty based on their own road test experiences, such as
the velocity of the vehicles in the traffic that the AV is likely to
run into, road condition, and AV braking distance. We discuss more
detailed implications of the reward and penalty
in \Section~\ref{sec:score-implication}. Appendix~\ref{sec:appendix}
provides more detailed discussion of these variables.

No clear boundary is defined between safe and
  unsafe in various AV safety models~\cite{shalev2017formal,
    Nister:2019:SFF}. As such, the safety score does not intend to give a
threshold of safe versus unsafe either. Instead, it implies the level of
safety -- higher safety score (could be either positive or
negative) indicates safer AV system design.

\begin{figure}[tb!]
  \centering
  \includegraphics[width=0.8\linewidth]{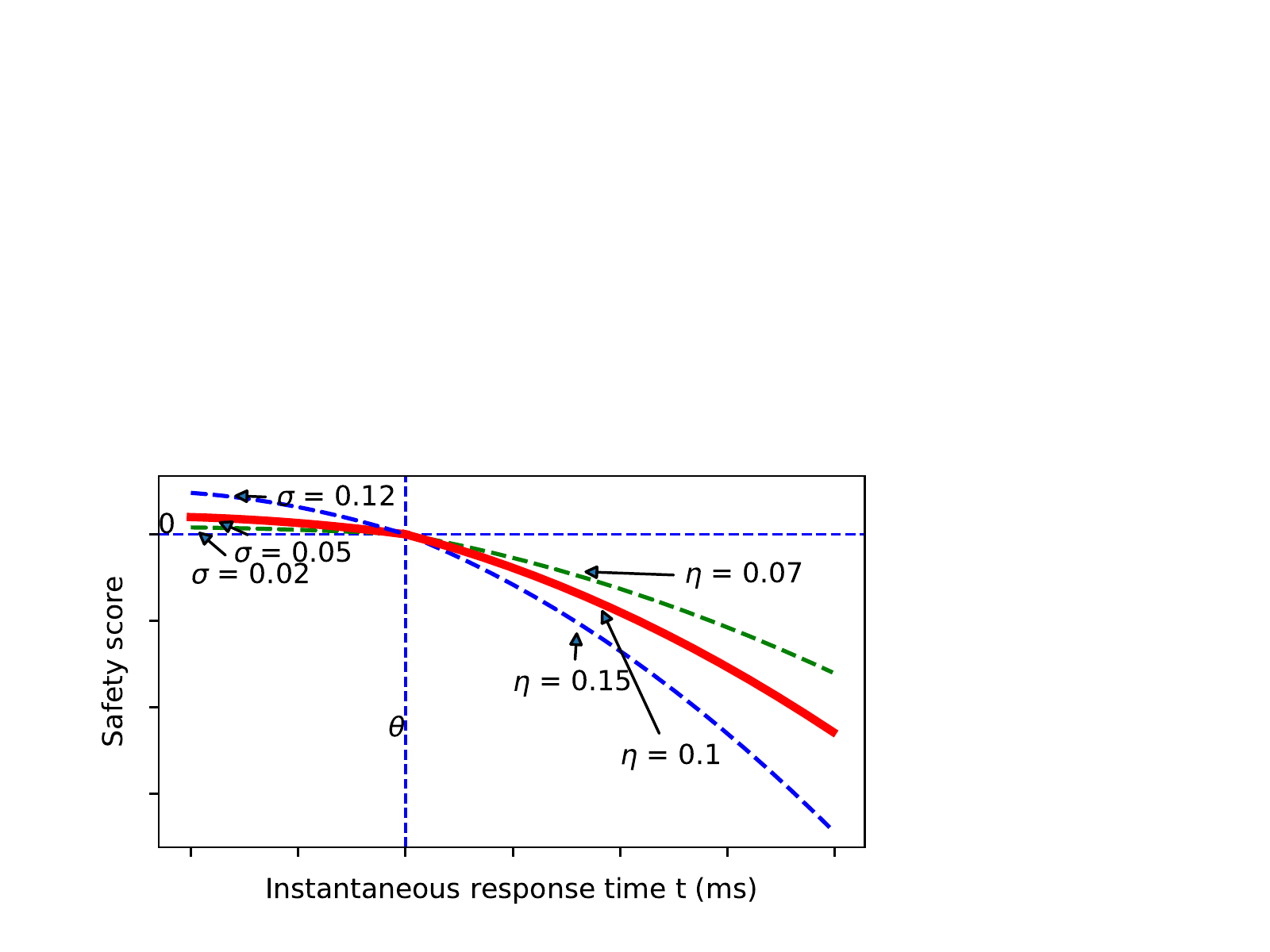}
    \vspace{-10pt}
  \caption{The non-linear relationship between safety score and
    instantaneous response time.}
  \label{fig:safetyscore-curve}
  \vspace{-10pt}
\end{figure}

\subsection{Implications to Computing System Design}
\label{sec:score-implication}

Based on our practical experience, safety score provides the following
key implications on the safety-aware AV system design.

\vspace{2pt}
\noindent\textbf{Nonlinearity.} The safety score is a non-linear
function of instantaneous response time $t$. Safety-aware computing
system design approach highly relies on the shape of the curve.
Figure~\ref{fig:safetyscore-curve} lists a few examples of safety
score curves as a function of $t$. With a given set of external
factors, the shape of the curve is determined by response time window
$\theta$, safety reward $\sigma$, and safety penalty $\eta$. Based on
our road tests, the majority of safety emergency scenarios happen when
$t>\theta$. When $t<\theta$, further optimizing the instantaneous
response time does not significantly improve the level of safety.
Therefore, in practice the shape of the safety score curve appears
like the bold line ($\sigma$=0.05, $\eta$=0.1) shown in
Figure~\ref{fig:safetyscore-curve}, with a lower reward than penalty.
As such, computing system design needs to prioritize reducing the
instantaneous response time $t$, when $t>\theta$.

\vspace{2pt}
\noindent\textbf{Instantaneous latency.} As we discussed before, the
safety score is computed by instantaneous response time per sensor
frame. Therefore, it is infeasible to adopt traditional performance
metrics calculated based on statistical performance behaviors, such as
tail, average, and maximum and minimum
latency. \Section~\ref{sec:safety-score-vs-others} quantitatively
compares various metrics.

\vspace{2pt}
\noindent\textbf{Safety impact across various software modules.} Due
to the complex inter-dependency relationship and latency accumulation
effect of the software modules, the instantaneous latency of different
software module has different impact on safety score. This is modeled
by Equation~\ref{eq-effective-latency}, where each software module has
a disparate form of the contribution to the instantaneous AV system
response time.

\begin{figure*}[tb!]
  \centering
  \includegraphics[width=1.0\linewidth]{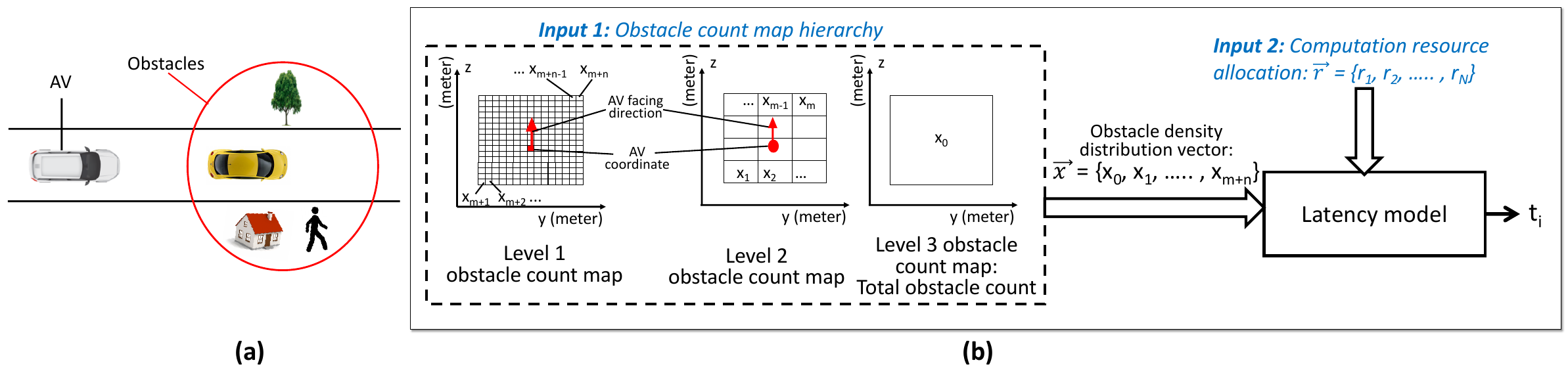}
    \vspace{-20pt}
  \caption{Overview of the proposed perception latency model. (a) Sources of perception latency. (b) Latency model working flow.}
  \label{fig:overview}
  \vspace{-10pt}
\end{figure*}

\section{Latency Model}
\label{sec:model}

The goal of our latency model is to estimate the instantaneous
response time in safety score, given obstacle distribution and
computation resource allocation. Our field study shows that perception
latency is a significant contributor to system response time
(\Section~\ref{sec:fieldstudy}). Therefore, we focus on modeling
perception latency.

Our model is built based on two observations: first, closer and denser
obstacles lead to higher perception latency
(\Section~\ref{sec:fieldstudy}); second, latency also depends on
computation resources allocated to run a software module\footnote{In
  our current AV system, such hardware resource allocation refers to
  whether executing a software module on CPU or GPU. Although our
  system does not adopt FPGA or other accelerators, our latency model
  applies to general heterogeneous architecture design with a variety
  of hardware resources}. Figure~\ref{fig:overview} shows an overview
of our latency model. The model input includes (1) obstacle density
distribution and (2) computation resource allocation of each software
module. The output is the estimated latency of each software module,
i.e., $t_i$ in Equation~\ref{eq-effective-latency}. Our latency model
estimates $t_i$ in two steps:

\squishlist

\item We first estimate a baseline latency ($\tau_i$), which is the
  latency of executing a software module on a baseline computation
  resource (e.g., CPU in our AV system), using a baseline latency
  model presented by Equation~\ref{eq-complexity}.

\item Then, we calculate $t_i$ with latency conversion ratios based on
  given resource allocation plan.

\squishend

\vspace{2pt}\noindent In the following, we use LiDAR perception as an
example to discuss our latency model in detail and can surely apply it
to other perception modules, such as camera, radar,
etc. \Section~\ref{sec:eval-coefficient} 
evaluates the accuracy of our latency model.

\subsection{Model Input}

\noindent\textbf{Obstacle density distribution vector.} We design an
obstacle count map to represent obstacle density distribution. The map
divides the sensor ROI into regular grids, with each grid storing the
number of obstacles in it. For example, our AV has a 64$m$ $\times$
50$m$ ROI with a 2$m\times$ 2$m$ grid size. This grid size captures
most moving objects on the road, e.g., vehicles and bicycles. But
large obstacles, which spread across multiple grids, are counted
multiple times. To express more accurate obstacle density, we adopt a
hierarchy of three levels of obstacle count maps with a grid size of
2$m\times$ 2$m$, 8$m\times$10$m$, and 64$m$ $\times$ 50$m$ (the whole
ROI), respectively. With the map hierarchy, we generate an obstacle
density distribution vector $\vec{x}$ as illustrated in
Figure~\ref{fig:overview}(b).

\vspace{2pt}
\noindent\textbf{Computation resource allocation vector.} We represent
computation resource allocation by a resource index vector $\vec{r}$,
where each element in the vector indicates the resource allocation of
a software module. For example, our current AV implementation only has
two resource allocation options, running on CPU (index=0) or GPU
(index=1). Then, $\vec{r}=[0, 0, 1, ..., 0]$ indicates that software
modules are executed on CPU, CPU, GPU, ..., CPU.

\subsection{Latency Model}

Perception software modules vary in complexity. We take into account
the diversity in the algorithm complexity and model the baseline
latency of each software module ($\tau_i$) by the following
equation,

\begin{equation} \label{eq-complexity}
\tau_i = \vec{a}\cdot(\vec{x}\odot \vec{x}) +
\vec{b}\cdot(\vec{x}\odot log(\vec{x})) + \vec{c}\cdot\vec{x} +
\vec{d}\cdot log(\vec{x}) + e
\end{equation}

Here, $x$ is a $m+n+1$ dimension obstacle density distribution vector,
where $m$ and $n$ are the grid counts in the finest and second-level
hierarchy, respectively. The operator $\odot$ is coefficient-wise
vector product, and $\cdot$ is the inner vector product.

This is a standard curve fitting problem. To solve for coefficients
$\vec{a}$, $\vec{b}$, $\vec{c}$, $\vec{d}$, and $e$ in the equation,
we perform linear regression on massive road testing data, which is
used to generate the model input data, i.e., obstacle density
distribution and corresponding baseline latency.

The final latency of each perception module then can be estimated by

\begin{equation} \label{eq-latency}
t_i = \tau_i\nu(r_i)\\
\end{equation}

Here, $\nu(r_i)$ is the ratio of executing the same
module on a different computation resource than the baseline. This
ratio is determined by exhaustive search of running a software module
with various computation resource allocation configurations.

\section{AV System Resource Management}
\label{sec:impl}

We demonstrate an example of utilizing our safety score and latency
model to guide the computing system design, by designing a
heterogeneous computation resource management scheme. The goal of
resource management is to determine the computation hardware resource
allocation of each software module and the priority of software module
execution to optimize safety score.

\textbf{Overview.} Our resource management has two phases, resource
planning and resource scheduling. During the \emph{planning phase}, we
perform an exhaustive search of computation resource allocation and
priority options of executing software modules in various obstacle
distributions, to determine the options that maximize safety score
given each obstacle distribution. This yields a resource management
plan for each obstacle distribution. During the \emph{scheduling
  phase}, we match the current obstacle distribution with resource
management plans and determine the plan to use. To ensure sufficient
resources to be scheduled for each software module, our AV system
maintains redundant computation resources. Due to the large searching
space, resource planning usually takes a long time. Therefore, we
perform the planning phase offline, while the AV system only performs
online scheduling. To further accelerate the online scheduling,
offline planning groups obstacle distributions associated with the
same resource management plan to clusters. As such, the scheduling
phase only needs to match clusters.

\subsection{Offline Planning}\label{sec:offline}

The offline planning phase analyzes the massive data collected by road
tests to create a set of resource management plans with various
obstacle distributions. A resource management plan includes (i) the
computation resources allocated for software module execution (CPU or
GPU in our current AV system design) and (ii) the priority of software
modules, which is used to determine which module gets executed first,
when multiple choices exist.

Algorithm~1 illustrates the offline planning procedure. For each field
study sample, which comprises instantaneous computing system response
time and the latency of each software module for an obstacle
distribution, we perform an exhaustive search in the space of resource
management plans. We categorize the samples, which achieve the highest
safety score for the same resource management plan, in the same
cluster. Based on our field study, samples in the same cluster have
similar (1) obstacle density distribution and (2) perception timeout
pattern, which is the number of continuous incidents where the
perception latency exceeds the sensor sampling interval (e.g., a LiDAR
sampling interval is 100ms). Therefore, we concatenate these two as
the feature, and compute the feature vector of each cluster by
averaging the feature of each sample in the cluster.

This phase only need to be done once, whenever the AV adapts to a new
location or physical system. In practice, we found that the clustering
and the associated resource management plans are stable over time,
even though users can also periodically run the offline phase to
re-calibrate.

\begin{algorithm}
\label{algorithm}
%\small
\caption{Offline planning.}
\begin{algorithmic}[1]
\State $Input: Obstacle\ density\ distribution\ vector\ S \in \mathbb{R}^{N \times m}$
\State $Input: Timeout\ pattern\ K \in \mathbb{R}^{N \times m}$
\State $Input:\ Dependency\ graph\ G$
\State $Input:\ Computation\ resource\ management\ plans\ P$
\State $Input:\ Safety~ score \ function\ L $ 
\State $Output: \ Category\ features\ F \in \mathbb{R}^{k \times 2m}$
\State $Init:\ Category\ Dictionary\ D $
\For {$i = 1 \to N$}
\State $P^*_{i}\ =\ \operatorname*{argmax}_{P_j} L(G, P_{j}, S_{i}) $
\State $append\ S_i\ \oplus \ K_i \ to\ D[P^*_i] $
\EndFor
\For {$j = 1 \to k$}
\For {$q = 1 \to m$}
\State $F_{j, q} \leftarrow mean(D[P_j]_q) $
\EndFor
\EndFor
\end{algorithmic}
\end{algorithm}

\subsection{Online Monitoring and Scheduling}
\label{sec:online}

\begin{figure}[tb!]
  \centering
  \includegraphics[width=\linewidth]{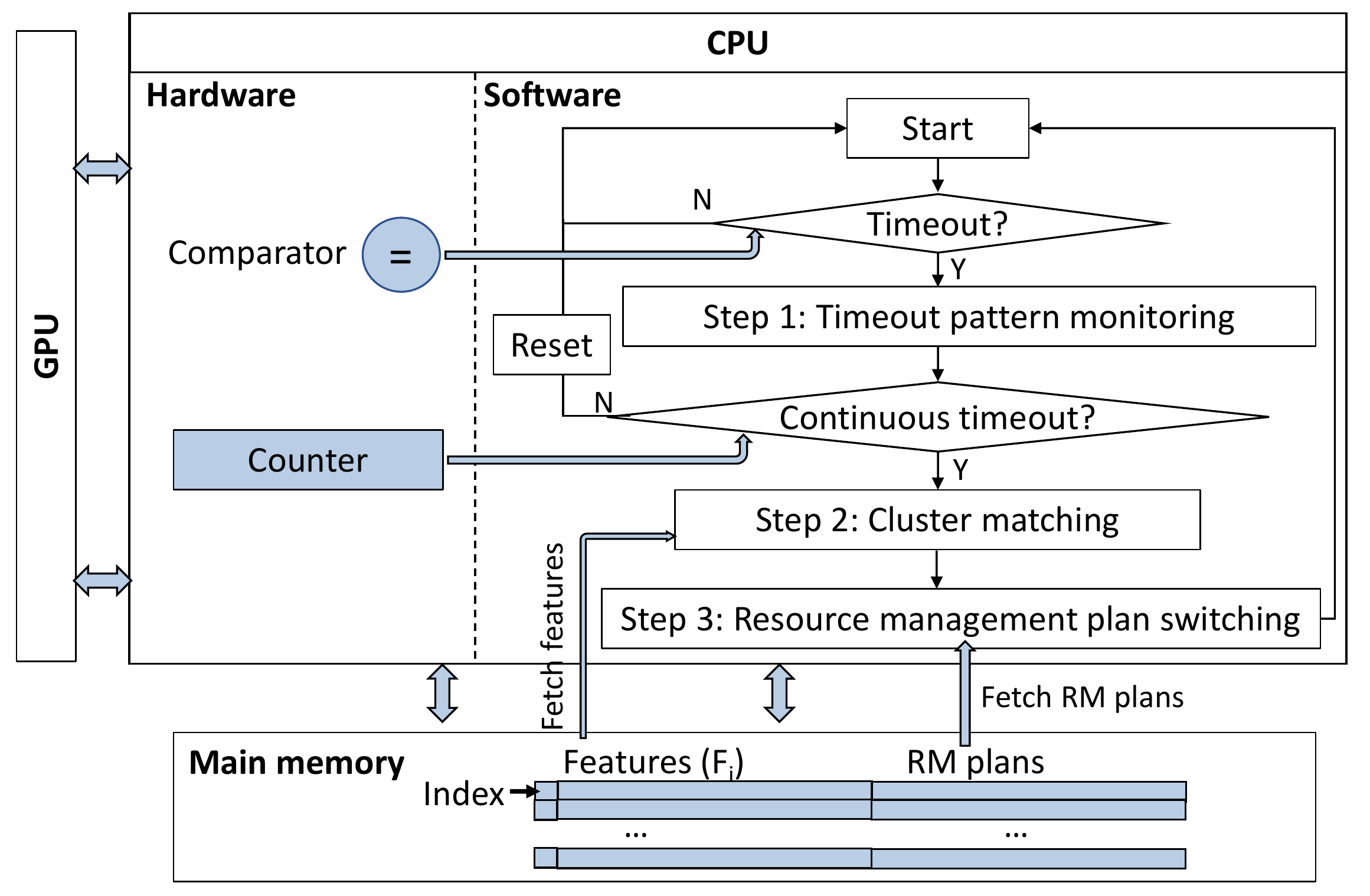}
    \vspace{-10pt}
  \caption{AV computing system design that performs safety-aware
    online monitoring and scheduling.}
  \label{fig:architecture}
  \vspace{-5pt}
\end{figure}

The online phase adopts a software/hardware co-design approach to (1)
monitor the computing system execution and (2) match the resource
management plans. Figure~\ref{fig:architecture} shows a system design
overview. When the AV is driving, the computing system continuously
(1) monitors instantaneous system response time and the pattern of
perception timeout and (ii) determines when and which resource
management plan to adopt.

A loop of three steps implements this: (Step-1) perception timeout
pattern monitoring, (Step-2) cooperative cluster matching, and
(Step-3) plan switching. The online phase triggers a hardware-based
perception timeout pattern monitoring, whenever one of the safety
critical perception software modules has a timeout (e.g. LiDAR); the
monitoring hardware counts the number of continuous timeouts across
the subsequent LiDAR sampling intervals to identify whether timeout
happens accidentally or continuously, e.g., over a user-defined $h$
continuous timeouts. Based on our field study, LiDAR perception
timeout happens either accidentally for a few times or continuously
for more than 100 frames. Therefore, we set $h$ to be 100. If
continuous timeout happens, the online phase performs the next two
steps based on the following equations:

\begin{equation}\label{eq-online1}\vspace{-10pt}
index = \operatorname*{argmin}_{i}\lVert F_{current} -  F_i \rVert
\end{equation}

\begin{equation}\label{eq-online2}
p_i = D[index]
\end{equation}

where $F_{current} = S_{current} \oplus K_{current}$. This is the
current feature vector, which is the concatenation of current obstacle
density distribution vector $S_{current}$ and current timeout pattern
$K_{current}$; $\oplus$ is the concatenation operator. $F_i$ is the
feature vector of the $i_{th}$ cluster, which consists of the obstacle
density distribution vector and timeout pattern of the $i_{th}$
cluster. Assuming road condition and traffic pattern do not
dramatically change within a certain amount of time, the online
resource management software will scan the features of all clusters
and identify a matching cluster, by finding the cluster with the
closest feature as the current one.

\vspace{2pt}
\noindent\textbf{Hardware and software implementation.} To meet the
real-time requirement of online resource management, we implement
Step-1 in CPU hardware by adding a set of comparators and counters as
shown in Figure~\ref{fig:architecture}. The comparators track the
timeouts; the counters accumulate the number of continuous timeouts.
In our AV system, an 8-bit counter is sufficient to capture the
continuous timeout patterns. Step-2 and Step-3 are triggered once
Step-1 detects continuous timeouts. As a result, these Step-2 and
Step-3 are performed infrequently. As such, we implement the primary
functionality of these two steps by software as a load
balancer~\cite{Mittal:2015:SCH, Lee:2013:TCC} in CPU; it acts as a
reverse proxy of software module scheduling across CPU and GPU. The
user-level implementation imposes negligible performance overhead
demonstrated by our experiment results, due to the infrequent cluster
matching and resource management plan switching. In our AV system, the
software modules that can execute on both CPU and GPU are implemented
and compiled with both versions. We store the indices generated during
Step-2 in CPU main memory. Step-3 reads the memory to identify the
resource management plan with the current feature.
	 
\section{Evaluation}
\label{sec:evaluation}

Our evaluation employs two simulators. First, to evaluate safety
score, we use an in-house AV simulator, which replays and analyzes
road testing data\footnote{The simulator is also used by the company
  to analyze routine road testing data.}. %and probability of collisions. 
  Second, we use an in-house computer architecture
simulator to evaluate the effectiveness of the online resource
monitoring and scheduling in our resource management scheme. We
compare three computing system architecture configurations and
resource management policies:

\squishlist

\item \textbf{CPU} -- Executing all software modules on CPU;
\item \textbf{CPU+GPU} -- The native heterogeneous AV system
  design. It employs tail latency to determine the priority of
  software module execution in resource management, with the GPU to
  accelerate all deep learning software modules.
\item \textbf{Resource management} -- Heterogeneous AV computing
  system design with our resource management scheme. The results take
  into account the performance overhead of the online monitoring and
  scheduling phase.

\squishend

\subsection{The Need for Safety Score}
\label{sec:safety-score-vs-others}

Our safety-score, which is rigorously derived from the RSS model,
measures the level of safety of AV system design. Furthermore, our AV
simulator results show that the level of safety identified by our
safety score is in line with the level of safety based on the
probability of collisions.

\begin{figure}[tb!]
  \centering
  \includegraphics[width=\linewidth]{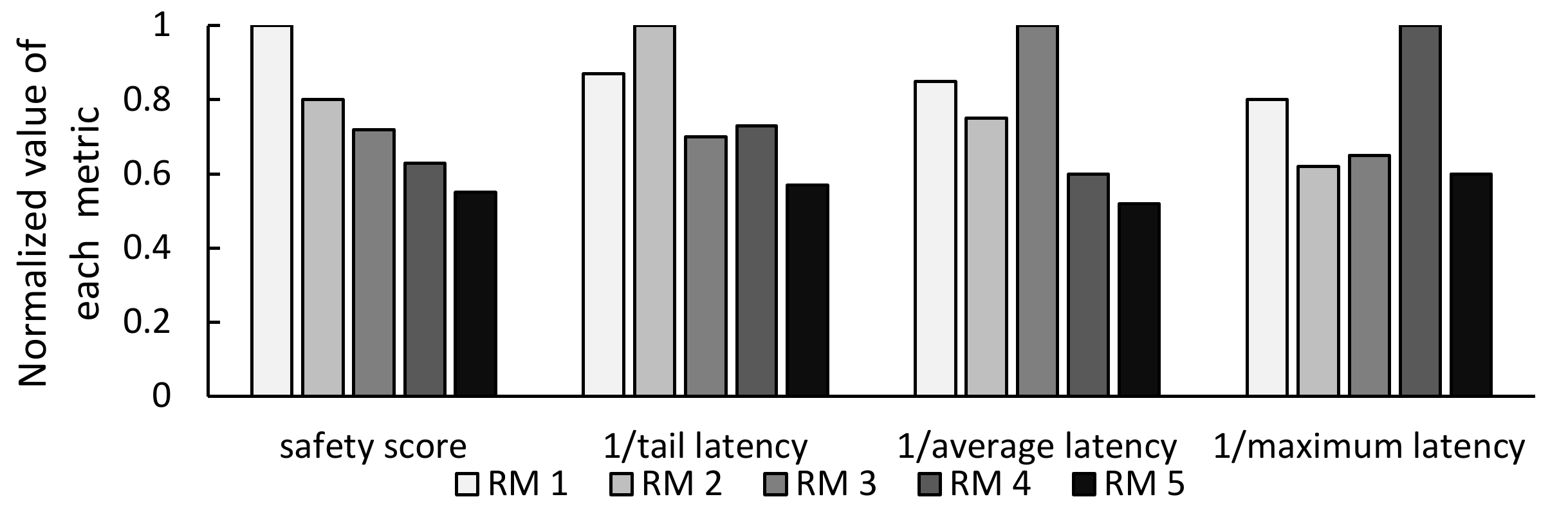}
  \vspace{-10pt}
  \caption{Resource management (RM) guided by various metrics (y-axis
    is normalized to the maximum of each metric).}
  \label{fig:framework-comparison}
  \vspace{-10pt}
\end{figure}

To demonstrate that other traditional latency metrics do not
accurately indicate the level of safety, we compare computing system
resource management results guided by safety score, 95th percentile
tail latency (99th percentile tail latency yields similar results),
average latency, and maximum latency. We perform an exhaustive search
of resource management plans for software modules. Due to space
limitation, Figure~\ref{fig:framework-comparison} shows an example of
five resource management plans. But the following observations hold
with our exhaustive search. For fair comparison, we present the
results obtained under the same set of obstacle distributions.

The results (the higher the better) show that safety-score-guided
resource management leads to entirely different results than other
metrics. For example, the safety score indicates that we should choose
\emph{RM-1}, whereas \emph{RM-2}, \emph{RM-3}, and \emph{RM-4} leads
to the best tail latency, average latency, and maximum latency,
respectively. Therefore, guiding the computing system design by these
traditional latency metrics can be misleading; it is critical to adopt
safety score.

\subsection{Latency Model Analysis}
\label{sec:eval-coefficient}

\begin{figure}[tb!]
  \centering
  \includegraphics[width=\linewidth]{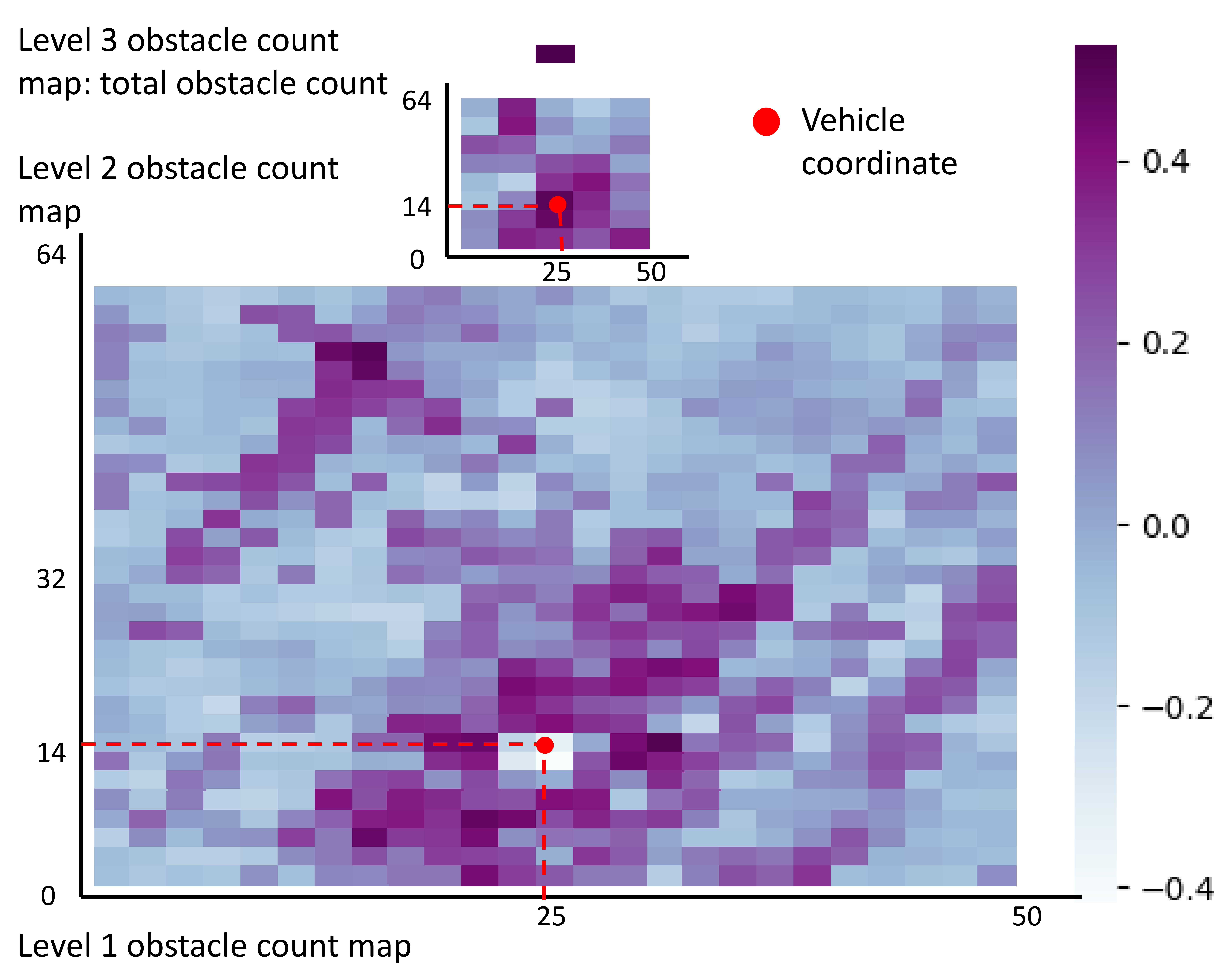}
   \vspace{-10pt}
  \caption{Heat map of Pearson correlation coefficient for obstacle count map hierarchy.}
  \label{fig:pearson}
  %\vspace{-10pt}
\end{figure}

\noindent\textbf{Pearson correlation coefficient.} We adopt the
Pearson correlation coefficient~\cite{benesty2009pearson} to evaluate
the correlation between LiDAR perception latency and obstacle density
distribution. Figure~\ref{fig:pearson} depicts a heat map of Pearson
correlation coefficients distributed in an obstacle count map
hierarchy, based on real driving scenarios during our field study.
Instead of the count of obstacles, each grid in
Figure~\ref{fig:pearson} stores a Pearson correlation coefficient. The
coefficient measures the correlation between obstacle density and
LiDAR perception latency: the higher the coefficient, the larger the
correlation between the two. Figure~\ref{fig:pearson} uses colors to
express Pearson correlation coefficient values: the darker the color,
the larger the value. As such, in the areas with darker colors, LiDAR
perception latency is more susceptible to obstacle density. We make
three observations from Figure~\ref{fig:pearson}. First, LiDAR
perception latency is more susceptible to nearby obstacles than those
far away. This is in line with our field study observation. Second,
both the top left and top right areas have high Pearson coefficient.
This may be caused by heavy horizontal traffic through an intersection
during rush hours. Finally, LiDAR perception latency is more sensitive
to the density than the distribution of obstacles.

\vspace{3pt}
\noindent\textbf{Model accuracy.} We use mean squared error
(MSE)~\cite{tuchler2002minimum} (the lower the better) to evaluate the
accuracy of our perception latency model quantitatively. We
investigate our latency model under a variety of driving scenarios,
such as morning and afternoon, rush hours and non-rush hours, and
local roads and express ways. The proposed latency model consistently
provides high accuracy with an average MSE as low as $1.7 \times
10^{-4}$.

\begin{figure}[tb!]
  \centering
  \includegraphics[width=1.0\linewidth]{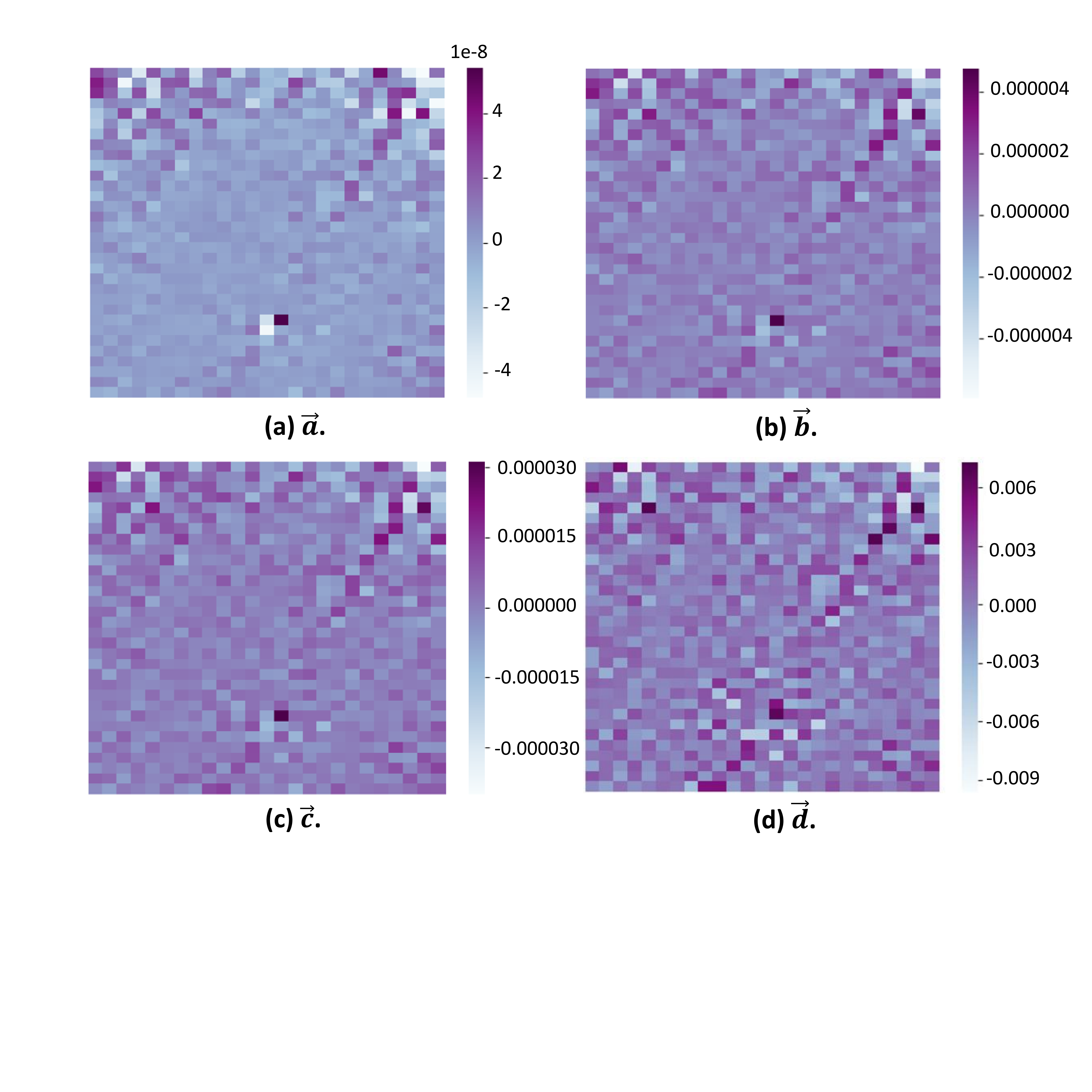}
   \vspace{-10pt}
  \caption{Heat maps of perception latency model coefficients
    $\vec{a}$, $\vec{b}$, $\vec{c}$, and $\vec{d}$.}
  \label{fig:heatmap}
  %\vspace{-10pt}
\end{figure}

\vspace{3pt}
\noindent\textbf{Model coefficient.} We borrow the above heat map
methodology to analyze the coefficients in our perception latency
model. Figure~\ref{fig:heatmap} shows the coefficients obtained based
on our field study data. We make two observations. First, most
coefficients are non-zero. Second, coefficients of different terms in
Equation~\ref{eq-complexity} have a significant variation on values.
The coefficients of lower-order terms have larger values than those of
higher-order terms. For instance, $\vec{a}$
(Figure~\ref{fig:heatmap}(a)), which are the coefficients of the
highest order term $x^2$ in Equation~\ref{eq-complexity}, has the
smallest values at a scale of $10^{-8}$.

\begin{figure}[tb!]
  \centering
  \includegraphics[width=\linewidth]{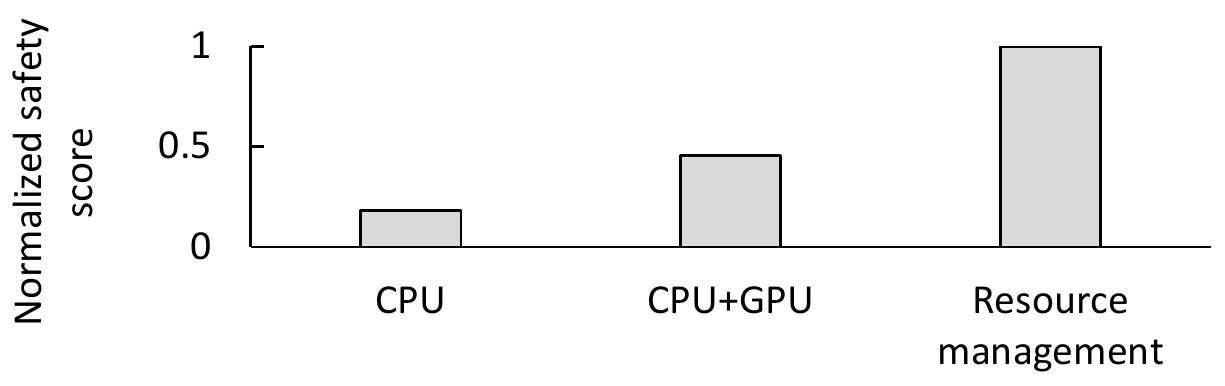}
   \vspace{-10pt}
  \caption{Safety score with various architecture configurations.}
  \label{fig:safetyscore}
  %\vspace{-5pt}
\end{figure}

\begin{figure}[tb!]
  \centering
  \includegraphics[width=\linewidth]{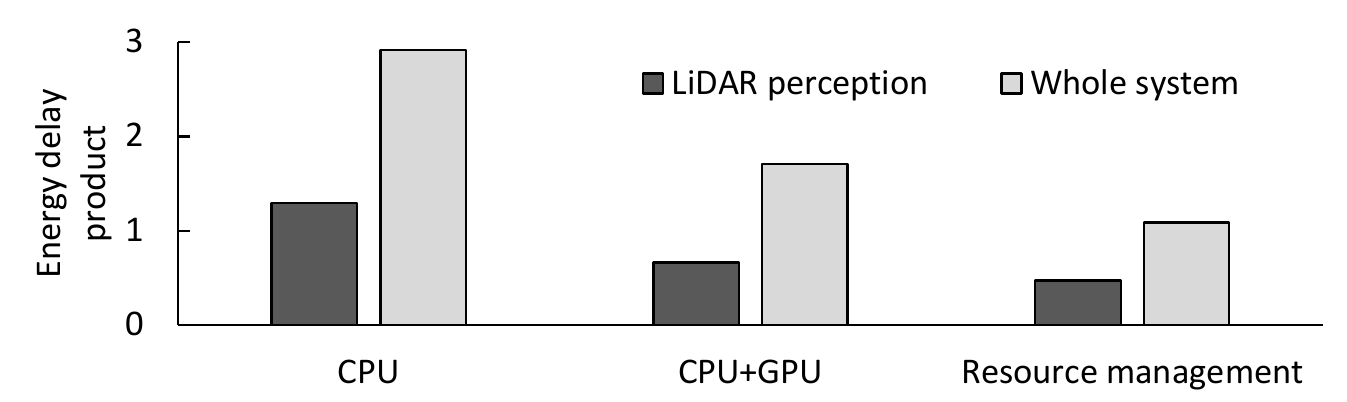}
    \vspace{-10pt}
  \caption{%\todo{split to two separate figures. These figures are too small and we have extra space now to have separate figures.}
  Energy efficiency of various AV computing system designs.} 
  \label{fig:EDP}
  %\vspace{-10pt}
\end{figure}

\begin{figure}[tb!]
  \centering
  \includegraphics[width=\linewidth]{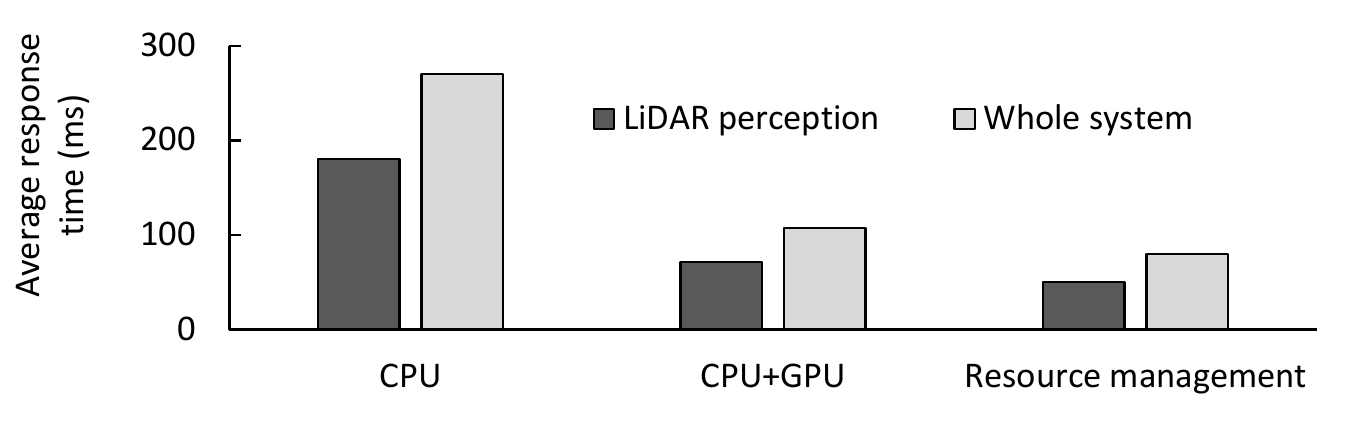}
    \vspace{-10pt}
  \caption{Average response time of various computing system designs.}
  \label{fig:average-latency}
  %\vspace{-10pt}
\end{figure}

\subsection{Resource Management Analysis}

Figure~\ref{fig:safetyscore} compares the safety score of three
different AV system architecture configurations and resource
management policies. Compared with \emph{CPU} and \emph{CPU+GPU},
\emph{Resource Management} achieves the highest safety score, which is
$1.3\times$ and $4.6\times$ higher than \emph{CPU+GPU} and \emph{CPU},
respectively.

Beyond improving safety, \emph{Resource Management} also improves
computing system performance and energy efficiency.
Figure~\ref{fig:EDP} compares the energy-delay product (EDP) of
various configurations. \emph{Resource Management} leads to the
highest EDP in both LiDAR perception alone and the whole computing
system. Our hardware modifications to implement online resource
management only impose less than 1\% of total system energy
consumption. Figure~\ref{fig:average-latency} compares the average
response time of various configurations. Compared with \emph{CPU} and
GPU+CPU, \emph{Resource management} reduces the response time averaged
throughout our field study data by $2.4\times$ and 35\%, respectively;
LiDAR perception latency is reduced by $2.6\times$ and 45\% on
average.

\section{Related Work}
\label{sec:related-work}

To our knowledge, this is the first paper to propose a safety metric
and latency model used for guiding AV computing systems and
architecture design, based on a large-scale field study of real
industry Level-4 AVs. In this section, we discuss related works on AV
nominal safety policies, safety-aware AV system design, and
heterogeneous computing system design.

\vspace{3pt}
\noindent\textbf{AV nominal safety policies.} Two recent studies,
RSS~\cite{shalev2017formal} and SFF~\cite{Nister:2019:SFF},
investigated AV nominal safety policies. Both studies define the
high-level concept of AV nominal safety. Our paper is in line with
this high-level safety concept. Unfortunately, neither of the studies
investigates safety requirement on computing systems and architecture
design. SFF~\cite{Nister:2019:SFF} focuses on policies that ensure
safe driving decisions, e.g., braking, acceleration, and steering,
under given limitations, such as perception visibility and AV system
response time. SFF guides the computing system software algorithm
design to enforce safety policies. But the study does not provide
policies on timely driving decision making, which is one of the
essential goals of computing systems and architecture design.
RSS~\cite{shalev2017formal} develops a set of general rules to ensure
collision avoidance, by enforcing that the AV keeps a safe distance
with surrounding vehicles. To this end, the safety rules treat the AV
and surrounding vehicles as objects with given velocity and
acceleration, without implications on the requirement of AV internal
systems (including the computing system). Our study focuses on the
nominal safety requirements of AV computing systems and architecture
design.

\vspace{3pt}
\noindent\textbf{Safety-aware AV system design.} Most previous
safety-aware AV system designs focus on developing software module
algorithms~\cite{ozguner2007systems, kunz2015autonomous,
  furda2011enabling, furda2009towards, laugier2011probabilistic,
  althoff2009model, helldin2013presenting, chu2012local}. For
instance, recent works improve the latency and precision of LiDAR- and
camera-based perception, by leveraging video sensing, laser
rangefinders, light-stripe range-finder to monitor all around the
autonomous driving vehicle to improve safety with a map-based fusion
system~\cite{aufrere2003perception, ozguner2007systems,
  kunz2015autonomous}. Several studies enhance the quality of driving
decisions by optimizing perception and planning software
algorithms~\cite{furda2011enabling, furda2009towards}. Previous works
also investigate ways of effectively responding to uncertainty at
real-time~\cite{laugier2011probabilistic, helldin2013presenting,
  althoff2009model} and path planning to avoid static
obstacles~\cite{chu2012local}.

\vspace{3pt}
\noindent\textbf{AV computing system design.} Most previous studies on
AV systems and architecture design focus on accelerator design and
energy efficiency improvement of deep learning or computer vision
software execution~\cite{lin2018-autonomous-driving,
  jo2015development}. The performance and energy optimizations are
guided by standard performance metrics. Safety requirement is not
adequately considered in most of these studies. As our evaluation
demonstrates, safety-aware computing system design leads to different
architecture and systems designs compared with guided by traditional
design metrics. However, our metric and latency model can also be used
to guide deep learning and computer vision accelerator design in AV
systems.

\vspace{3pt}
\noindent\textbf{Heterogeneous computing system design.} AV computing
systems adopt sophisticated heterogeneous architectures used in
real-time scenarios. Prior works on real-time heterogeneous computing
systems~\cite{song2018mis, dang2017anomaly, chen2015towards} focus on
embedded processors, rather than sophisticated architectures. Previous
studies that investigate server-grade CPU and GPU processor based
systems focus on distributed and data center
applications~\cite{Klimovic:2018:SHC, krauter2002taxonomy,
  beloglazov2010energy, maheswaran1999dynamic, czajkowski1998resource,
  beloglazov2012energy}; in such cases, scalability is a more critical
issue than single-system resource limitation. Furthermore, neither
type of heterogeneous system use cases has the level of safety
requirements as autonomous driving.
  
\section{Conclusion}
\label{sec:conclusion}

In summary, this paper proposes a safety score and perception latency
model to guide safety-aware AV computing system design. Furthermore,
we elaborated detailed AV computing system architecture design and
workloads, as well as discussing a set of implications of safety-aware
AV system design based on a field study with industrial AVs, and
demonstrate the use of our safety score and latency model with a
safety-aware heterogeneous computation resource management scheme. As
high automation vehicles are still at the early stages of development,
we believe our work only scratched the surface of safety-aware AV
system design. Substantial follow-up work is required to enhance the
computing system design towards safe autonomous driving.

%%%%%%% -- PAPER CONTENT ENDS -- %%%%%%%%

\newpage

%%%%%%%%% -- BIB STYLE AND FILE -- %%%%%%%%
\bibliographystyle{IEEEtranS}
\bibliography{ref}
%%%%%%%%%%%%%%%%%%%%%%%%%%%%%%%%%%%%

\begin{appendix}
\section{Safety Score Formulation}\label{sec:appendix}

Responsibility-Sensitive Safety (RSS)~\cite{shalev2017formal}
formalizes safety as assuring five common-sense rules, which AV
driving policy should obey. Specifically, the policy will choose safe
actions under various circumstances. The action is safe, if (1) given
the current driving state and the actions and (2) when the surrounding
vehicles are acting adversely but reasonably, a collision can be
avoided.

More formally, a given driving state $\{v, a_{max, +}, a_{min,-}\}$ of
the AV and the driving state $\{v', a'_{max, +}, a'_{min, -}, a'_{max,
  -}\}$ of the vehicle that might have collision with the AV (referred
to as ``obstacle-of-attention''), together determine the minimum
distance $d_{min}$ to avoid collision by Equation~\ref{eq:d_min}:

\begin{align}
    d_{min} = v t + \frac{1}{2}a_{max, +}t^2 + \frac{(v+t a_{max, +})^2}{2 a_{min, -}} \nonumber\\
    + m(v' t' + \frac{1}{2}a'_{max, +} t^{'2}) \nonumber\\ 
    + 
    n \frac{(v' + t' a'_{max, +})^2}{2a'_{*, -}} + d_{\mu}\label{eq:d_min}
\end{align}

Here, $v$ and $v'$ are velocities of the AV and the
obstacle-of-attention, respectively; $a_{min/max, +/-}$ is the
minimum/maximum acceleration of speeding up($+$) or slowing down($-$).
Let $t$ and $t'$ be the response times of the AV and the
obstacle-of-attention. All variables are non-negative: velocities and
accelerations only represent the absolute value.

Parameters $m$ and $n$ are used to represent whether the two vehicles
are driving in the opposite or the same directions. The values are
scenario-dependent. For example, $(m=1, n=1)$ indicates that the two
vehicles, the AV and the obstacle-of-attention, are driving
towards each other, either along the lane direction or perpendicular
to the lane direction. The reasonable conditions are the
obstacle-of-attention takes some response time $t'$ before it applies
a brake; the most adverse condition is to use the minimum slowing
down acceleration $a'_{*,-} = a'_{min,-}$. When the two vehicles are
driving towards the same direction, which is represented by $(m=0,
n=-1)$, the most adverse condition is that the obstacle-of-attention
instantly brakes ($t'=0$) with highest possible acceleration $a'_{*,-}
= a'_{max,-}$.

Parameters $t', a'_{*, +/-}, v', a_{*, +/-}, d_{\mu}$ depend on
various environmental conditions and user's safety preference. In
practice, under bad weather or road conditions, we may adopt lower
$v'$ and $a'_{max, +/-}$ than normal conditions, because vehicles tend
to drive slower and apply cautious speed changes. $d_{\mu}$ is the
minimum distance between the two vehicles, when they perform safe
actions~\cite{shalev2017formal} and full stop. Users may specify
larger $d_{\mu}$ to allow a larger safety margin or smaller $d_{\mu}$
to achieve higher driving flexibility. Higher $a_{*. +/-}$ enables the
AV to react more agile, but may lead to the worse passenger
experience. At any time, the AV speed $v$ is known. The AV sensors
detect the speed of the obstacle-of-attention. Collapsing all the
parameters and grouping by the only variable $t$ gives a quadratic
form as Equation~\ref{eq:quad}. Notice that $\alpha$ and $\beta$ are
non-negative.

\begin{align}
    & d_{min} = \alpha t^2 + \beta t + \gamma , \nonumber\\
    &where ~
    \alpha = \frac{1}{2}a_{max, +} +  \frac{ a_{max, +}^2}{2 a_{min, -}}, ~
    \beta = v + \frac{va_{max, +}}{a_{min, -}}, ~ \gamma = \frac{v^2}{2a_{min, -}}\label{eq:quad}
\end{align}

% Except $\{v, a_{max, +}, a_{min,-}, \rho\}$, other parameters in the Equation~\ref{eq:d_min} are exogenous. $a'_{max/min, +/-}$ depend on both physical capability of vehicles and soundness/usefulness tradeoff: if the autonomous driving car assumes the acceleration its surrounding vehicles is very high, its driving policy will be very cautious and driving might fall impossible. Moreover, in different driving scenarios, the $m$ and $\rho'$ that will lead to the most adverse condition are summarized in Table~\ref{tab:cond}. \todo{JZ: $\mu$ is user determined safe distance threshold?}
% \begin{table}[!ht]
%     \centering
%     \label{tab:cond}
%     \begin{tabular}{|c|c|c|c|c|}
%     \hline
%          Scenario & m & $t'$ & $a'_{*, -}$ & $\mu$\\
%          \hline
%          Safe longitudinal distance (Same direction) & 0 & 0 & $a'_{max, -}$ & 0 \\
%          Safe longitudinal distance (Opposite direction) & -1 & $t'$ & $a'_{min, -}$ & 0 \\
%          Safe lateral distance & -1 & $t'$ & $a'_{min, -}$ & $\mu$ \\
%     \hline
%     \end{tabular}
%     \caption{Caption}
%     \label{tab:my_label}
% \end{table}

According to RSS~\cite{shalev2017formal}, if the current distance $d$
is larger than $d_{min}$, the safety criteria is met. Without loss of
generosity, a safety score function can be defined as
Equation~\ref{eq:lin}. $\sigma$ is the reward specified by users to
achieve extra margin from minimal safety distance; $\eta$ is the
penalty of unit distance less than the minimum safety distance.

\begin{equation}
\label{eq:lin}
    \textrm{Safety Score}=\left\{
                \begin{array}{ll}
                 \sigma (d - d_{min}), if d > d_{min}\\
                 \eta (d - d_{min}), elsewhere
                \end{array}
              \right.
\end{equation}

We employ Equation~\ref{eq:quad} to bridge the safety score with
computing system response time. First, the current distance $d$
between the AV and the obstacle-of-attention will determine the
response time window $\theta$ that will avoid the collision. The
response time window $\theta$ is the time that the AV would take to
hit the obstacle-of-attention, if both vehicles keep moving at the
same velocity and acceleration -- or more formally, if both vehicles
conform the Duty of Care, i.e. an individual should exercise
"reasonable care" while performing actions that might harm the safety
of others~\cite{shalev2017formal}. For instance, the Duty of Care
allows the front car to perform break, such that the distance to the
rear car is reduced, but prohibits the front car to backup, when
followed by another car. We determine $\theta$ by solving the
following function:

\begin{equation}
d=\alpha \theta ^2 + \beta \theta + \gamma
\end{equation}

\begin{equation}
\label{eq:fin}
    \textrm{Safety Score}=\left\{
                \begin{array}{ll}
                 \sigma [\alpha(\theta ^2 - t^2) + \beta(\theta - t)], if \theta > t\\
                 \eta [\alpha(\theta ^2 - t^2) + \beta(\theta - t)] , elsewhere
                \end{array}
              \right.
\end{equation}

\begin{figure}[t!]
    \center
    \includegraphics[width=0.8\linewidth]{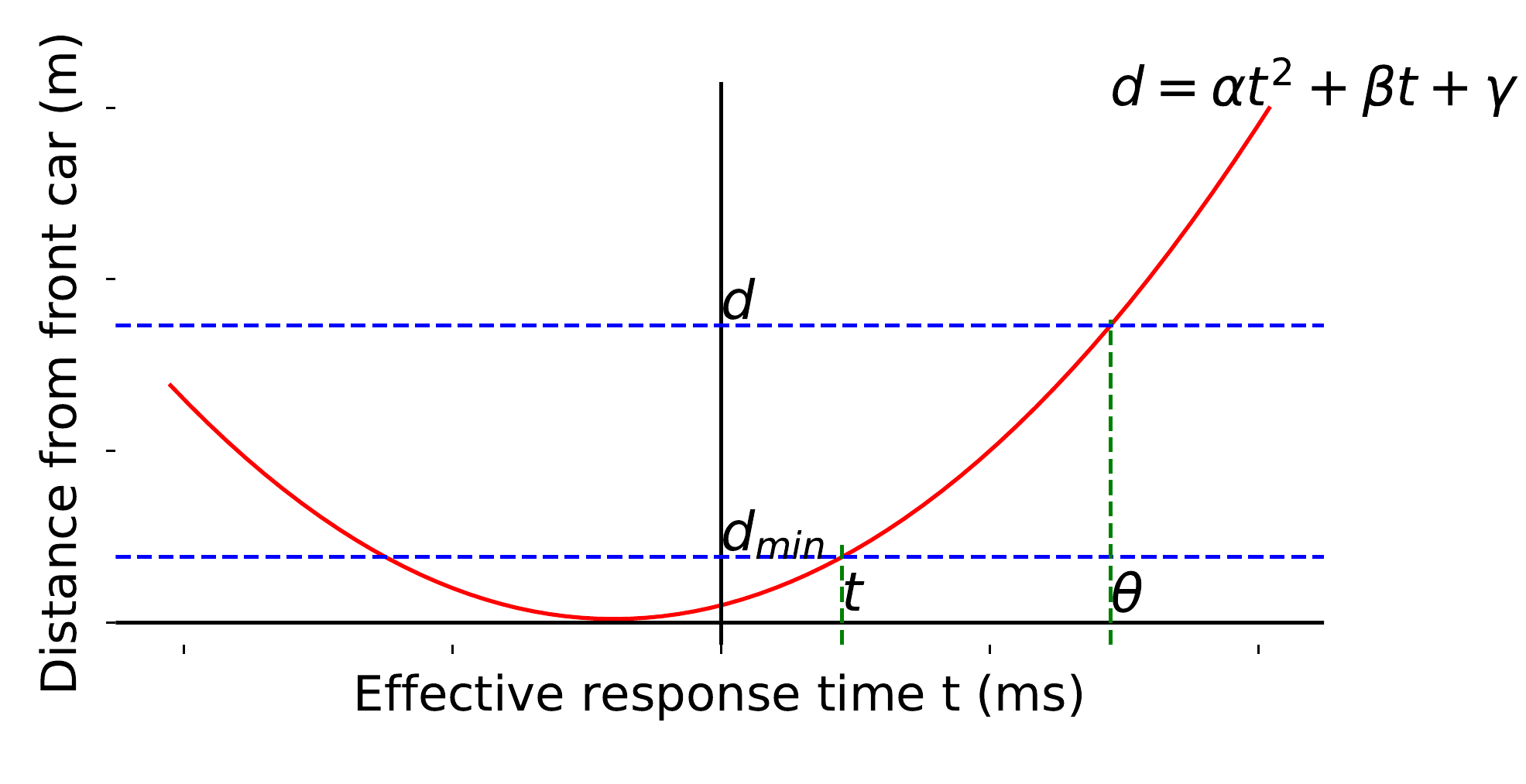}
\caption{Illustration of d-$\theta$ and t-$d_{min}$.}
\label{fig:ap1}
\end{figure}

Let $t$ be the AV computing system response time and determine
$d_{min}=\alpha t^2 + \beta t + \gamma$. Figure~\ref{fig:ap1} depicts
the relationship between Equation~\ref{eq:quad} with the current
distance $d$ and instantaneous computing system response time $t$.
Then the safety score function can be written as
Equation~\ref{eq:fin}.

\end{appendix}

\end{document}